\documentclass[journal]{IEEEtran}
\usepackage{amsmath,amsfonts}
\usepackage{algorithmic}
\usepackage{algorithm}
\usepackage{array}
\usepackage[caption=false,font=normalsize,labelfont=sf,textfont=sf]{subfig}
\usepackage{textcomp}
\usepackage{stfloats}
\usepackage{url}
\usepackage{verbatim}
\usepackage{graphicx}
\usepackage[nospace,nocompress]{cite}
\usepackage{pifont}
\usepackage{xspace}
\def\onedot{\ifx\@let@token.\else.\null\fi\xspace}
\def\eg{\emph{e.g}\onedot}

\def\ie{\emph{i.e}\onedot}

\def\Vec#1{{\boldsymbol{#1}}}
\def\Mat#1{{\boldsymbol{#1}}}

\usepackage{amsthm}

\usepackage{booktabs}
\usepackage{multirow}
\usepackage[table]{xcolor}

\begin{document}

\definecolor{lwqgreen}{rgb}{0.1412, 0.5451, 0.0157}

\title{SRasP: Self-Reorientation Adversarial Style Perturbation for Cross-Domain Few-Shot Learning}

\author{Wenqian Li, Pengfei~Fang$^*$, Hui~Xue$^*$,~\IEEEmembership{Member,~IEEE}
\IEEEcompsocitemizethanks{\IEEEcompsocthanksitem  W. Li, P. Fang and H. Xue are with the School of Computer Science and Engineering, Southeast University, Nanjing 210096, China and the Key Laboratory of New Generation Artificial Intelligence Technology and Its Interdisciplinary Applications (Southeast University), Ministry of Education, Nanjing 211189, China 
        (E-mail: \{wenqianli.li, fangpengfei, hxue\}@seu.edu.cn).}
        \thanks{$^*$ Corresponding author}
        }

\markboth{IEEE TRANSACTIONS ON PATTERN ANALYSIS AND MACHINE INTELLIGENCE}%
{Shell \MakeLowercase{\textit{et al.}}: A Sample Article Using IEEEtran.cls for IEEE Journals}


\maketitle

\begin{abstract}
Cross-Domain Few-Shot Learning (CD-FSL) aims to transfer knowledge from a seen source domain to unseen target domains, serving as a key benchmark for evaluating the robustness and transferability of models. Existing style-based perturbation methods mitigate domain shift but often suffer from gradient instability and convergence to sharp minima. To address these limitations, we propose a novel crop–global style perturbation network, termed \underline{S}elf-\underline{R}eorientation \underline{A}dversarial \underline{S}tyle \underline{P}erturbation (SRasP). Specifically, SRasP leverages global semantic guidance to identify incoherent crops, followed by reorienting and aggregating the style gradients of these crops with the global style gradients within one image. Furthermore, we propose a novel multi-objective optimization function to maximize visual discrepancy while enforcing semantic consistency among global, crop, and adversarial features. Applying the stabilized perturbations during training encourages convergence toward flatter and more transferable solutions, improving generalization to unseen domains. Extensive experiments are conducted on multiple CD-FSL benchmarks, demonstrating consistent improvements over state-of-the-art methods.
\end{abstract}

\begin{IEEEkeywords}
Cross-Domain, Few-Shot Learning, Adversarial Style Perturbation, Image Classification.
\end{IEEEkeywords}

\section{Introduction}\label{sec:introduction}
\IEEEPARstart{D}{eep} learning models have achieved remarkable success in visual recognition tasks when trained on large-scale labeled datasets. Nevertheless, acquiring sufficient and reliable annotations is often impractical in many real-world scenarios, such as rare disease diagnosis. Few-Shot Learning (FSL) has therefore emerged as an effective paradigm that enables models to recognize novel categories from only a few labeled samples per class~\cite{triantafillou2019meta},~\cite{feng2024wave},~\cite{baik2023learning}. In practical deployment, an additional and more critical challenge arises from domain shifts between training and testing environments, which can severely degrade recognition performance. This challenge motivates the study of Cross-Domain Few-Shot Learning (CD-FSL), which aims to generalize knowledge learned from source domains to diverse unseen target domains~\cite{11164927},~\cite{feng2024transferring}. Among various CD-FSL settings, Single-Source CD-FSL represents a particularly realistic yet challenging scenario, where transferable representations must be learned from only one source domain.

\begin{figure}[!t]
\centering
\includegraphics[width=\linewidth]{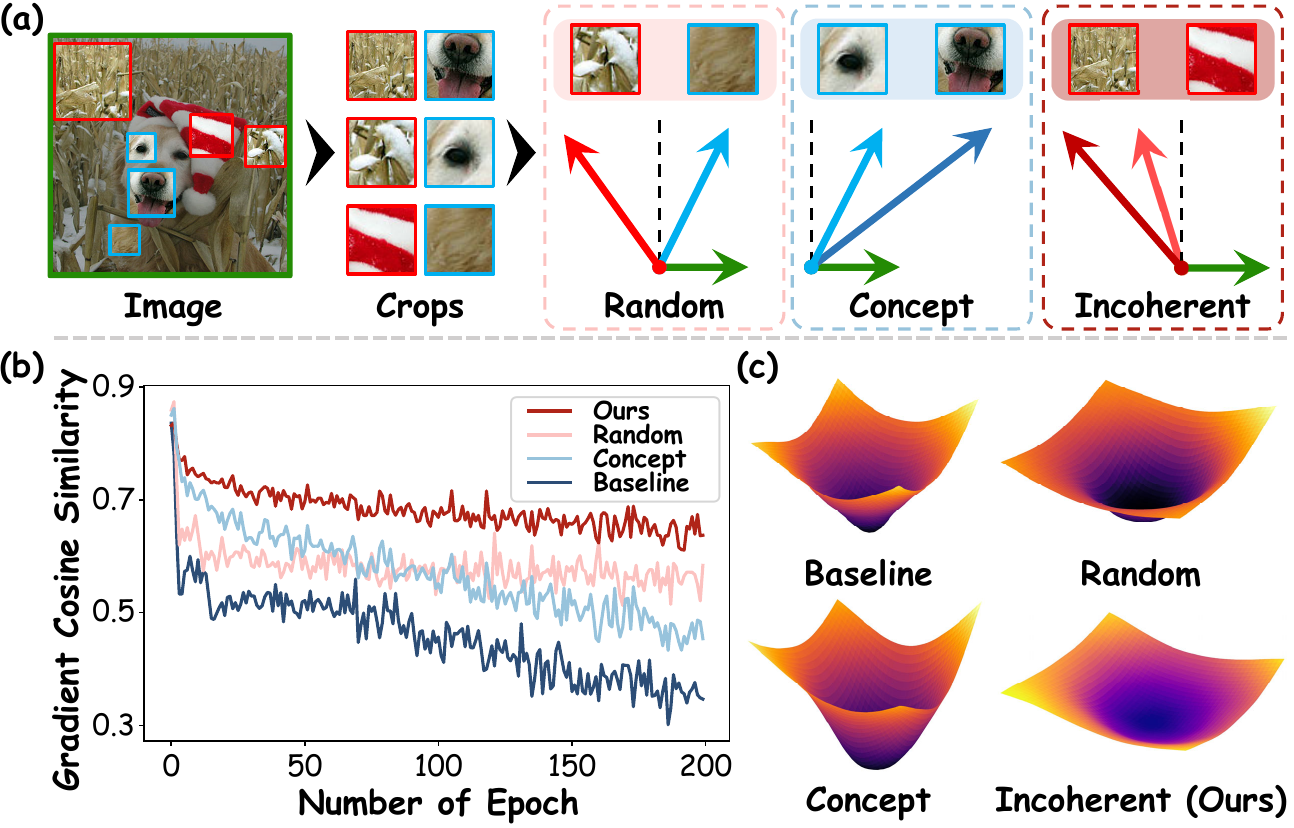}
\caption{(a) Given an input image, multiple local crops are extracted. The training process applies one of three crop selection strategies, including random-crop selection, concept-crop selection and incoherent-crop selection, each leading to different gradient directions during optimization. (b) The gradient cosine similarity across training epochs shows that the proposed method maintains consistently higher stability compared with other perturbation methods, indicating a more reliable update trajectory. (c) Loss-surface visualizations further demonstrate that our incoherent-crop perturbation drives the model toward flatter and more generalizable minima.}
\label{fig:intro}
\end{figure}

To mitigate domain shifts under this restrictive setting, recent studies have explored style-based perturbation as effective techniques to encourage the learning of domain-agnostic knowledge from a single source domain~\cite{kim2024randomized},~\cite{zhong2022adversarial}. These approaches are motivated by the observation that image styles (\eg mean and standard deviation) encode key domain-specific characteristics~\cite{zhou2020domain}. By perturbing such attributes, existing methods aim to suppress domain bias and improve generalization to unseen domains~\cite{feng2023genes},~\cite{xie2024kind}. Despite promising empirical results, current style-based CD-FSL methods often exhibit unstable optimization behaviors. Large inter-domain discrepancies combined with adversarial style perturbations lead to highly varying optimization paths, making training susceptible to gradient instability and convergence to sharp minima.

A key factor underlying this instability lies in the heterogeneous composition of images. As illustrated in Fig.~\ref{fig:intro}(a), an image can be decomposed into multiple local crops with different semantic relevance. Some crops capture discriminative foreground content and contribute positively to correct classification, referred to as \emph{concept crops}. In contrast, other crops are dominated by background textures or incidental visual patterns that are weakly related to semantic concepts, referred to as \emph{incoherent crops}. These heterogeneous regions naturally give rise to different crop selection strategies, including random-crop selection, concept-crop selection, and incoherent-crop selection, each inducing distinct optimization directions when incorporated into adversarial style perturbation.

Existing style-based CD-FSL methods perform perturbation solely on the global image, thereby overlooking the heterogeneous contributions of local crops. As shown in Fig.~\ref{fig:intro}(b) (Baseline), such global-only perturbation results in pronounced gradient instability, manifested as oscillatory and inconsistent update trajectories~\cite{fu2023styleadv}. Our preliminary work~\cite{li2025svasp}, as shown in Fig.~\ref{fig:intro}(b) (Random), partially alleviates this issue by randomly selecting and aggregating crop style gradients, yielding improved stability. Building upon this observation, we further conduct a systematic investigation of crop mining and gradient aggregation strategies. As shown in Fig.~\ref{fig:intro}(b) (Incoherent), our analysis reveals that reorienting and aggregating style gradients of incoherent crops is substantially more effective in stabilizing optimization, achieving consistently smoother update trajectories and improved robustness.

\begin{figure*}[!t]
\centering
\includegraphics[width=\linewidth]{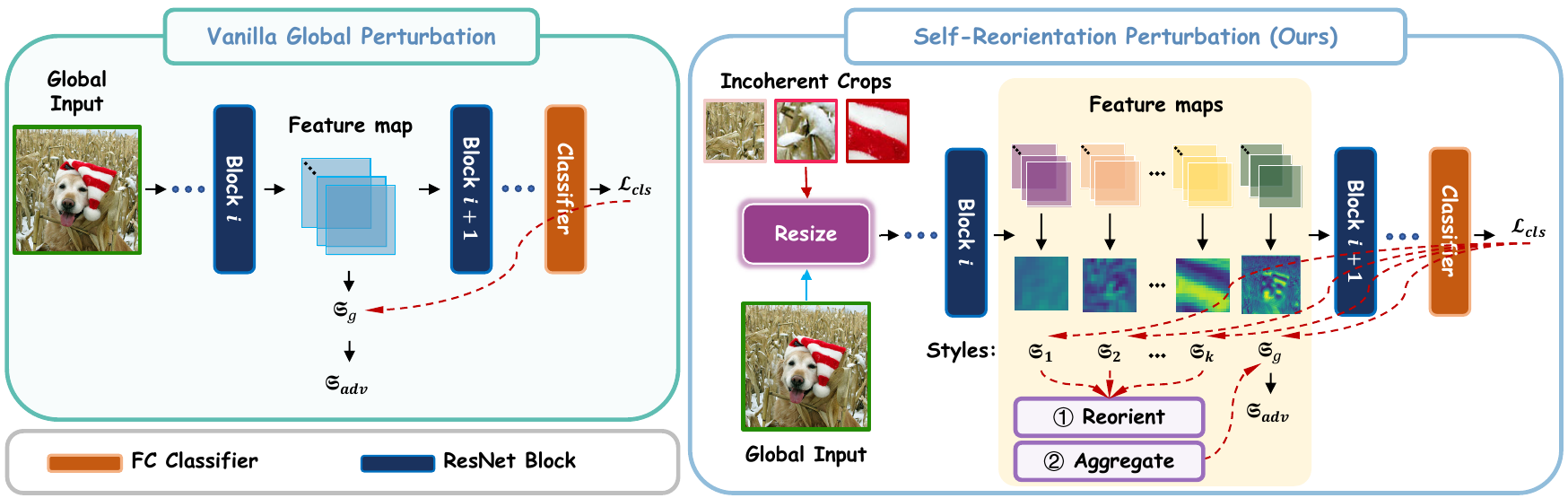}
\caption{Comparison between Vanilla Global Perturbation and our Self-Reorientation Perturbation. \textit{Left}: In the vanilla approach, a global style perturbation is directly applied to the feature map of the entire input. \textit{Right}: In our Self-Reorientation Perturbation, the input image is first divided into incoherent crops. These crop style gradients are then reoriented and aggregated with the global style gradient.}
\label{fig:intro2}
\end{figure*}

Motivated by these findings, we argue that incoherent crops should not be simply discarded but instead be systematically exploited. Although incoherent crops contain spurious and domain-specific appearance patterns, they provide challenging style variations that are essential for learning robust and transferable representations. The key challenge lies in leveraging such challenging perturbations without allowing their noisy gradients to dominate or misguide the global optimization trajectory.

To address this challenge, we propose Self-Reorientation Adversarial Style Perturbation (\textbf{SRasP}), a novel crop–global perturbation network designed to stabilize adversarial optimization in CD-FSL. As illustrated in Fig.~\ref{fig:intro2}, instead of perturbing only the global image, SRasP jointly considers the global image and multiple incoherent crops to synthesize adversarial styles, thereby explicitly modeling heterogeneous style variations within an image. Specifically, our method employs a two-level optimization structure consisting of outer and inner iterations. In each outer iteration, incoherent crops are identified under the guidance of global semantic supervision and subsequently employed in the inner iterations. During each inner iteration, the style gradients derived from these incoherent crops are iteratively reoriented and aggregated with the global style gradients. This self-reorientation mechanism suppresses conflicting gradient components while preserving hard yet semantically meaningful perturbations, effectively aligning incoherent-induced updates with the global semantic descent direction. Furthermore, we propose a novel Consistency–Discrepancy Triplet Objective (CDTO) to jointly promote visual diversity and semantic preservation. The proposed objective function maximizes visual discrepancy while enforcing semantic consistency between global, crop, and adversarial features, providing a robust supervisory signal. 

In contrast to the global-only perturbation strategy, SRasP substantially improves gradient stability during optimization, as evidenced in Fig.~\ref{fig:intro}(b). The mitigation of gradient oscillation enables the model to escape from poor sharp minima and to converge toward smoother and flatter minima, as further illustrated in Fig.~\ref{fig:intro}(c). To the best of our knowledge, this work constitutes one of the first systematic investigations into the impact of localized style gradients on model stability.

In summary, the contributions of this work are threefold:
\begin{itemize}
    \item We propose \textbf{SRasP}, a novel network that reorients and aggregates incoherent crop style gradients with the global style gradients within an image, a process we refer to as self-reorientation. This design stabilizes adversarial optimization and escapes sharp minima.
    \item We propose a new objective function, named Consistency–Discrepancy Triplet Objective (CDTO) to maximize visual discrepancy and enforce semantic consistency between global, crop, and adversarial features, providing a robust supervisory signal for CD-FSL.
    \item We conduct extensive experiments on multiple benchmark datasets and validate the effectiveness of our modules. The quantitative results show that our proposed SRasP significantly outperforms existing state-of-the-art (SOTA) methods.
\end{itemize}

\section{Related Work} \label{sec:related}
\subsection{Cross-Domain Few-Shot Learning}
First introduced in FWT~\cite{tseng2020cross}, Cross-Domain Few-Shot Learning (CD-FSL) aims to train a feature extractor on source domains that can generalize to novel domains~\cite{li2022cross},~\cite{oh2022understanding}. Unlike conventional Few-Shot Learning (FSL)~\cite{xue2025towards},~\cite{zhang2025reliable},~\cite{guo2025understanding}, CD-FSL suffers from severe domain shifts caused by discrepancies in style, background, and object appearance between source and target domains~\cite{das2021confess},~\cite{hu2022switch}. These discrepancies often lead to significant performance degradation. To address this issue, existing methods primarily focus on learning domain-invariant features through feature alignment~\cite{zhao2023dual},~\cite{wang2023mmt} or extending meta-learning with domain-aware adaptation strategies and regularization~\cite{xumemrein},~\cite{kang2025task},~\cite{yang2024leveraging}. Another line of work leverages style perturbation techniques to expose models to a broader range of visual variations, thereby enhancing robustness to unseen domains~\cite{yirevisiting},~\cite{zhou2024meta},~\cite{li2026hap},~\cite{sreenivas2023similar}. Nevertheless, CD-FSL remains highly challenging due to extremely limited target-domain supervision and large domain discrepancies.

\subsection{Region Mining and Local Feature Exploitation}
Beyond holistic representations, exploiting local discriminative regions has been shown to be particularly effective under limited-data regimes. Prior studies demonstrate that fine-grained object parts and informative patches often provide more stable class-discriminative cues than global embeddings, motivating region-aware and attention-based frameworks that emphasize salient spatial patterns while suppressing irrelevant background responses~\cite{lai2023spatialformer},~\cite{laiclustered},~\cite{chen2024featwalk}. Subsequent works further incorporate part-aware modeling and region-level mining to capture transferable local structures across categories \cite{jing2024fineclip}. This paradigm has recently been extended to CD-FSL. To mitigate these shifts, several methods adopt region mining and hard crop selection strategies to extract aligned foreground patches with improved robustness \cite{zhou2023revisiting},~\cite{zhuo2022tgdm},~\cite{ma2025reconstruction}. However, existing approaches primarily treat local crops as independent foreground enhancers and overlook the fact that incoherent regions can introduce biased gradients that distort global representations toward domain-specific patterns.

\subsection{Style-Based Adversarial Perturbation}
A growing body of work explores style-based augmentation and adversarial perturbation to enhance cross-domain generalization. Style transfer techniques, such as Adaptive Instance Normalization (AdaIN)~\cite{huang2017arbitrary}, modify appearance while preserving semantics, thereby simulating unseen target distributions. More recent approaches introduce adversarial perturbations in the style space~\cite{fu2023styleadv},~\cite{li2025svasp}, manipulating texture and feature statistics to regularize models against severe domain shifts. Despite their effectiveness, most existing methods rely on global style statistics and neglect inconsistencies between local and global styles, which can limit alignment quality. Our work advances this line by introducing a gradient-guided adversarial style perturbation mechanism that rectifies and integrates both global and local style gradients to synthesize coherent adversarial styles. Combined with a consistency–discrepancy objective, our framework enforces local stability while encouraging sufficient divergence under adversarial shifts, yielding more transferable representations.

\begin{figure*}
    \centering
    \includegraphics[width=\linewidth]{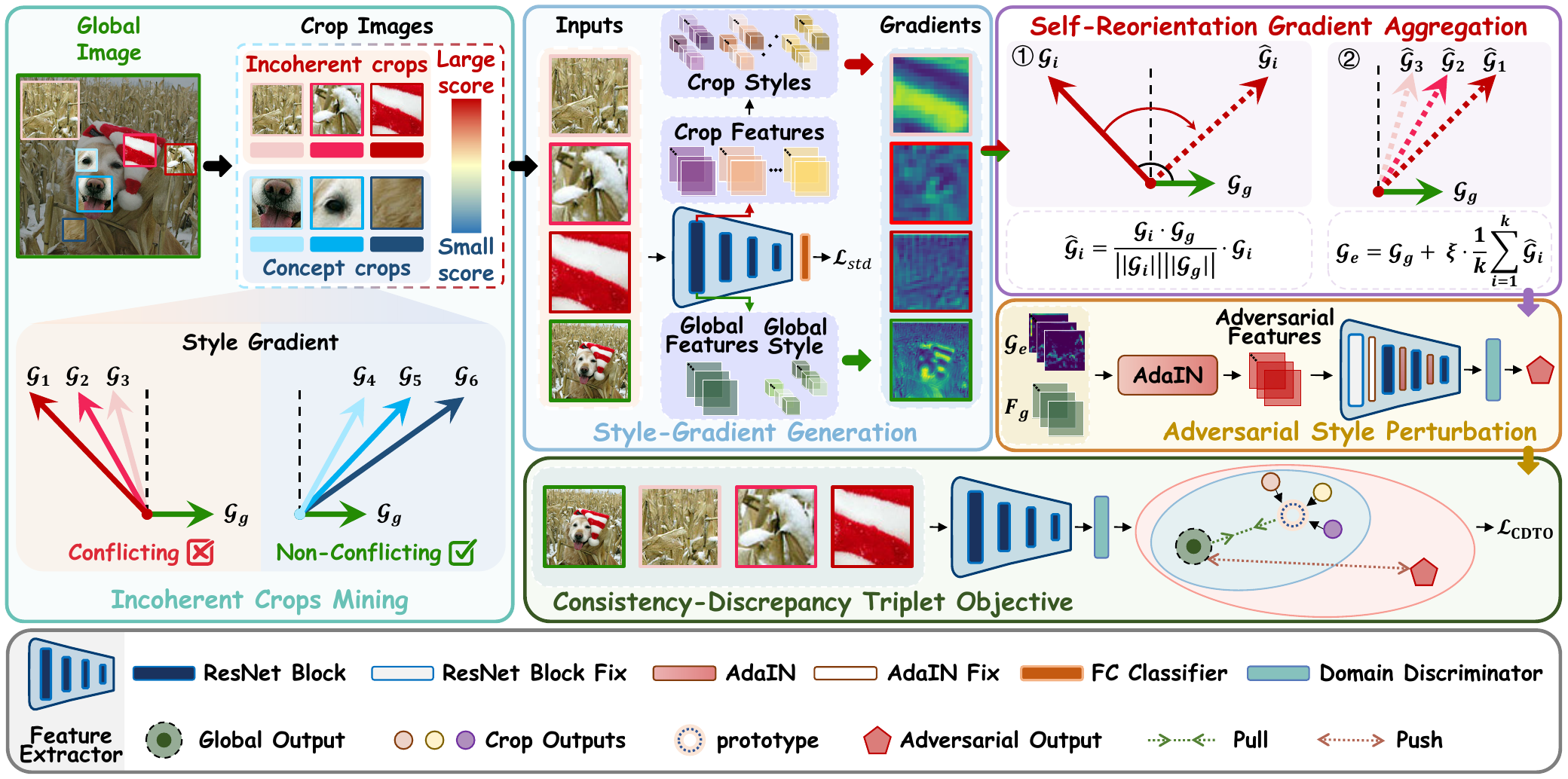}
    \caption{Overview of the proposed SRasP. SRasP first samples multiple localized crops from an input image and identifies incoherent regions that exhibit semantic inconsistency with the global content. Style gradients are extracted from both crop-level and global features, where incoherent crop gradients are reoriented toward the global semantic direction and aggregated through a self-reorientation gradient ensemble to resolve gradient conflicts. The resulting stable and semantically guided global style gradient is then used to synthesize hard yet meaningful adversarial style perturbations via AdaIN. In addition, a consistency–discrepancy triplet objective is employed to maximize visual style diversity while preserving semantic alignment among global, crop, and adversarial representations, enabling SRasP to generate stronger style variations and improve cross-domain generalization.}
    \label{fig:method}
\end{figure*}

\section{Methodology}
In this section, we first present a formal definition of the CD-FSL problem setting. Second, we introduce the proposed novel method SRasP, designed for CD-FSL. An overview of our method is depicted in Figure~\ref{fig:method}.

\subsection{Problem Formulation}
We investigate the Single Source CD-FSL setting, in which only a labeled source dataset $\mathcal{D}^s$ is available during training, whereas the target dataset $\mathcal{D}^t$ remains inaccessible. By definition of CD-FSL, the label spaces of the two domains are disjoint, \ie, $C(\mathcal{D}^s) \cap C(\mathcal{D}^t) = \emptyset$, and their underlying data distributions are also distinct, \ie, $P(\mathcal{D}^s) \neq P(\mathcal{D}^t)$, where $C(\cdot)$ and $P(\cdot)$ denote the categories and distributions of the source and target dataset, respectively. To mimic the Few-Shot Learning process, we employ the episode training paradigm. Specifically, in each episode, we construct an $N$-way $K$-shot task by sampling $N$ classes from $\mathcal{D}^s$, with $K$ labeled instances per class forming the support set $\mathcal{S} = \{\Mat{x}_i^s, {y}_i^s\}_{i=1}^{n_s}$, where $n_s = NK$. For the same $N$ classes, an additional $M$ unlabeled samples per class are drawn to establish the query set $\mathcal{Q} = \{\Mat{x}_i^q\}_{i=1}^{n_q}$, where $n_q = NM$. Hence, an episode can be denoted as $\mathcal{T} = (\mathcal{S}, \mathcal{Q})$, containing a total of $|\mathcal{T}| = N (K + M)$ samples. The ultimate objective is to learn a feature extractor and a classification head on the support set that can accurately predict the labels of the query set.
\subsection{SRasP}
\subsubsection{Overview} The proposed model contains a CNN/ViT feature extractor $E$, a global classification head $H_g$, a FSL relation classification head $H_r$ and a domain discriminator $H_{d}$ with learnable parameters $\theta_E$, $\theta_g$, $\theta_r$ and $\theta_d$, respectively.

SRasP consists of five modules: Incoherent Crops Mining module to identify incoherent crops whose visual styles are inconsistent with the global image semantics, Style-Gradient Generation module to extract style gradients of the global image and incoherent crops, Self-Reorientation Gradient Aggregation module to reorient and then aggregate the style gradients of the incoherent crops with the global style gradients, Adversarial Style Perturbation module to generate adversarial features by applying reoriented gradients, and Consistency-Discrepancy Triplet Objective module to maximize visual discrepancy while enforcing semantic consistency between global, crop, and adversarial representations.
    
\subsubsection{Incoherent Crops Mining}
To identify local regions that are least supportive of correct classification and most likely to introduce spurious style variations, we propose the incoherent crops mining strategy. The core intuition is that, under the same supervision signal, regions that incur larger classification loss tend to exhibit weaker semantic alignment and stronger visual inconsistency with the global image, making them a primary source of gradient instability during training. 

Specifically, given an input image $\Mat{x}$, we first generate a set of multi-scale crops $\{\Mat{c}_i\}_{i=1}^m$ using \textit{RandomResizedCrop} with diverse scale ranges, followed by standard data augmentations. Each crop is then forwarded through the feature extractor $E$ and the global classification head $H_g$, and its supervisory discrepancy score is quantified by the cross-entropy loss with respect to the ground-truth label $y$:

\begin{equation}
s_i = \mathcal{L}_{{CE}}\Big (H_g \big (E(\Mat{c}_i; \theta_E); \theta_g \big ) , y \Big ).
\end{equation}

From an optimization perspective, crops yielding lower loss values produce gradients that are well aligned with the global semantic direction, indicating that they capture discriminative and label-consistent visual cues. We therefore refer to such regions as \textit{concept crops}. In contrast, crops associated with higher loss values typically lack sufficient semantic evidence or contain background clutter and visually incoherent patterns. These regions tend to generate noisy or conflicting gradients and are thus identified as \textit{incoherent crops}. 

To explicitly emphasize these challenging regions, we rank all candidate crops according to their loss values and select the top-$k$ samples with the highest discrepancy to form the incoherent crop set:
\begin{equation}
    \mathcal{C}^{{inc}}
= \big \{ \Mat{c}_i | i \in \operatorname{topk}_{i}\big(s_i\big) \big \}.
\end{equation}
By focusing on incoherent crops, the proposed mining strategy deliberately exposes the model to hard and misleading local patterns that are often overlooked by conventional data augmentation. This design not only facilitates more informative and diverse style gradient generation in subsequent modules, but also provides a principled basis for stabilizing gradient aggregation under severe domain shifts.

 \subsubsection{Style-Gradient Generation}
In this paper, the styles of global and crop features are modeled as Gaussian distributions~\cite{liuncertainty} and learnable parameters which will be updated by adversarial training.  Specifically, for feature maps $\Mat{F} \in \mathbb{R}^{B \times C \times H \times W}$, where $B$, $C$, $H$ and $W$ denote the batch size, channel, height, width of the feature maps $\Mat{F}$, the specific formula for calculating the style $\mathfrak{S} = \{\Mat{\mu}, \Mat{\sigma}\}$ is:
\begin{equation}\label{eq:mu}
\Mat{\mu}=\frac{1}{HW}\sum_{h=1}^{H}\sum_{w=1}^{W}\Mat{F}_{B,C,h,w} ,
\end{equation}
\begin{equation}\label{eq:sigma}
\negthickspace \negthickspace \Mat{\sigma}=\sqrt{\frac{1}{HW}\sum_{h=1}^{H}\sum_{w=1}^{W}(\Mat{F}_{B,C,h,w}-\Mat{\mu})^{2} + \epsilon},
\end{equation}
where $\epsilon$ is a small value to avoid division by zero.

Instead of generating the adversarial style of all blocks' features at once, we use an iterative approach. Concretely, the embedding module $E$ has four blocks $B_1$, $B_2$, $B_3$, $B_4$, and style transformation is only performed on the first three blocks, as the shallow blocks produce more transferable features. For each block $B_j$ of the backbone $E$, we obtain the incoherent crop features and the global features $\mathbb{F}^j = \{\Mat{F}_{1}^j,\Mat{F}_{2}^j, \dots, \Mat{F}_{k}^j, \Mat{F}_g^j\}$. For each $\Mat{F}^j \in \mathbb{F}^j$, $\Mat{F}^j \in \mathbb{R}^{B \times C \times H \times W}$, $\Mat{F}^j$ is accumulated from block 1 to block $j-1$:
\begin{equation}
    \Mat{F}^j = \mathfrak{T}_j(\mathfrak{T}_{j-1}(\dots(\mathfrak{T}_1({\Mat{I}, \mathfrak{S}_{adv}^1}), \dots), \mathfrak{S}_{adv}^{j-1}), \mathfrak{S}_{adv}^j),
\end{equation}
where transferring features between block $j-1$ and block $j$ is formulated as:
\begin{equation}
    \mathfrak{T}_j({\Mat{F}^{j}, \mathfrak{S}_{adv}^j}) = \frac{B_j(\Mat{F}^{j-1}) - \Mat{\mu}_{F^j}}{\Mat{\sigma}_{F^j}} * \Mat{\sigma}_{adv}^j + \Mat{\mu}_{adv}^j,
\end{equation}
and the style $\mathfrak{S}_{F^j} = \{\Mat{\mu}_{F^j}, \Mat{\sigma}_{F^j}\}$ of $\Mat{F}^{j}$ is calculated by Eq.~\eqref{eq:mu} and \eqref{eq:sigma}.
Thus, we can get the styles of the feature maps of $B_j$ to form the style set $\mathbb{S} = \{ \mathfrak{S}_1^j, \mathfrak{S}_2^j, \dots, \mathfrak{S}_k^j, \mathfrak{S}_g^j\}$. Then, we continue to pass $\Mat{F}^j$ to the remainder of the backbone and the global classification head without performing any other operations and get the final prediction $\Vec{p}=H_{g}(B_4(\dots (B_{j+1}(\Mat{F}^j)));\theta_{g}), \Vec{p}\in \mathbb{R}^{B \times N_c}$, where $N_c$ denotes the total number of classes. Thus the total prediction set is $\mathbb{P} = \{\Vec{p}_{1}, \Vec{p}_{2}, \dots, \Vec{p}_{k}, \Vec{p}_{g}\}$. Therefore the classification loss can be written as:
\begin{equation}
    \mathcal{L}_{cls} = \mathcal{L}_{CE}(\Vec{p}_{g}, {y}) + \sum\nolimits_{i=1}^k\mathcal{L}_{CE}(\Vec{p}_{i}, {y}),
\end{equation}
where $\mathcal{L}_{CE}(\cdot, \cdot)$ denotes the cross-entropy loss. 

The sequel will compute the adversarial style of block $B_j$, we omit the subscript $j$ for readability and calculate the gradients of the mean $\Mat{\mu}$ and the std $\Mat{\sigma}$ by loss back propagation: 
\begin{equation}
\begin{aligned}
\mathbb{G}^{\mu} & = \{\boldsymbol{\mathcal{G}}^{\mu}_{1}, \boldsymbol{\mathcal{G}}^{\mu}_{2}, \dots, \boldsymbol{\mathcal{G}}^{\mu}_{k}, \boldsymbol{\mathcal{G}}^{\mu}_{g}\} \\
        & = \{\nabla_{\Mat{\mu}_{1}}\mathcal{L}_{cls}, \nabla_{\Mat{\mu}_{2}}\mathcal{L}_{cls}, \dots, \nabla_{\Mat{\mu}_{k}}\mathcal{L}_{cls}, \nabla_{\Mat{\mu}_{g}}\mathcal{L}_{cls}\},
\end{aligned}
\label{eq:g_mu}
\end{equation}
\begin{equation}
\begin{aligned}
\mathbb{G}^{\sigma} & = \{\boldsymbol{\mathcal{G}}^{\sigma}_{1}, \boldsymbol{\mathcal{G}}^{\sigma}_{2}, \dots, \boldsymbol{\mathcal{G}}^{\sigma}_{k}, \boldsymbol{\mathcal{G}}^{\sigma}_{g}\} \\
        & = \{\nabla_{\Mat{\sigma}_{1}}\mathcal{L}_{cls}, \nabla_{\Mat{\sigma}_{2}}\mathcal{L}_{cls}, \dots, \nabla_{\Mat{\sigma}_{k}}\mathcal{L}_{cls}, \nabla_{\Mat{\sigma}_{g}}\mathcal{L}_{cls}\},
\end{aligned}
\label{eq:g_sigma}
\end{equation}
Other blocks' style gradients can be generated likewise. 

\subsubsection{Self-Reorientation Gradient Aggregation}

While the aggregation of global and crop style gradients provides richer supervisory signals, a naive averaging of crop gradients may introduce inconsistency due to their heterogeneous optimization directions. To mitigate this issue, we propose the Self-Reorientation Gradient Aggregation mechanism, which reorients each crop style gradient according to its cosine similarity with the global style gradient before aggregation.

Formally, given the global style gradient \{$\boldsymbol{\mathcal{G}}^{\mu}_{g}$, $\boldsymbol{\mathcal{G}}^{\sigma}_{g}$\} and the gradients of the selected incoherent crops $\{\boldsymbol{\mathcal{G}}^{\mu}_{i}, \boldsymbol{\mathcal{G}}^{\sigma}_{i}\}_{i=1}^k$. The similarity between each crop style gradient and the global style gradient is computed as:
\begin{equation}
\gamma^{\mu}_i = \cos(\boldsymbol{\mathcal{G}}_{i}^{\mu}, \boldsymbol{\mathcal{G}}^{\mu}_{g}) = \frac{\langle \boldsymbol{\mathcal{G}}_{i}^{\mu}, \boldsymbol{\mathcal{G}}^{\mu}_{g} \rangle}{||\boldsymbol{\mathcal{G}}_{i}^{\mu}||_2 ||\boldsymbol{\mathcal{G}}^{\mu}_{g}||_2},
\label{eq:cosine_mu}
\end{equation}
\begin{equation}
\gamma^{\sigma}_i = \cos(\boldsymbol{\mathcal{G}}_{i}^{\sigma}, \boldsymbol{\mathcal{G}}^{\sigma}_{g}) = \frac{\langle \boldsymbol{\mathcal{G}}_{i}^{\sigma}, \boldsymbol{\mathcal{G}}^{\sigma}_{g} \rangle}{||\boldsymbol{\mathcal{G}}_{i}^{\sigma}||_2 ||\boldsymbol{\mathcal{G}}^{\sigma}_{g}||_2}.
\label{eq:cosine_sigma}
\end{equation}
We then rectify each crop gradient by projecting it along the global gradient direction:
\begin{equation}
\hat{\boldsymbol{\mathcal{G}}}_{i}^{\mu} = \gamma^{\mu}_i \cdot \boldsymbol{\mathcal{G}}_{i}^{\mu},
\quad
\hat{\boldsymbol{\mathcal{G}}}_{i}^{\sigma} = \gamma^{\sigma}_i \cdot \boldsymbol{\mathcal{G}}_{i}^{\sigma}.
\end{equation}
The reoriented crop gradients are then aggregated and normalized to get the crop style gradients $\mathbb{G}^c = \{\boldsymbol{\mathcal{G}}_{c}^\mu,\boldsymbol{\mathcal{G}}_{c}^\sigma\}$, where: 
\begin{equation}
    \boldsymbol{\mathcal{G}}_{c}^\mu = Norm(\frac{1}{k} (\hat{\boldsymbol{\mathcal{G}}}^{\mu}_{1} + \hat{\boldsymbol{\mathcal{G}}}^{\mu}_{2} + \dots + \hat{\boldsymbol{\mathcal{G}}}^{\mu}_{k})),
\label{eq:g_crop_mu}
\end{equation}
\begin{equation}
    \boldsymbol{\mathcal{G}}_{c}^\sigma = Norm(\frac{1}{k} (\hat{\boldsymbol{\mathcal{G}}}^{\sigma}_{1} + \hat{\boldsymbol{\mathcal{G}}}^{\sigma}_{2} + \dots + \hat{\boldsymbol{\mathcal{G}}}^{\sigma}_{k})).
\label{eq:g_crop_sigma}
\end{equation}

Subsequently, a decay factor $\xi$ is introduced to finally get the ensemble style gradients $\mathbb{G}^e = \{\boldsymbol{\mathcal{G}}_{e}^\mu,\boldsymbol{\mathcal{G}}_{e}^\sigma\}$, where:
\begin{equation}
    \boldsymbol{\mathcal{G}}_{e}^\mu = Norm(\boldsymbol{\mathcal{G}}_g^{\mu}) + \xi \odot \boldsymbol{\mathcal{G}}_{c}^\mu,
\label{eq:g_e_mu}
\end{equation}
\begin{equation}
    \boldsymbol{\mathcal{G}}_{e}^\sigma = Norm(\boldsymbol{\mathcal{G}}_g^{\sigma}) + \xi \odot \boldsymbol{\mathcal{G}}_{c}^\sigma.
\label{eq:g_e_sigma}
\end{equation}

\subsubsection{Adversarial Style Perturbation} We get the random initialized global styles $\Mat{\mathfrak{S}}_{init} = \{\Mat{\mu}_{init}, \Mat{\sigma}_{init}\}$ by adding Gaussian noise $\mathcal{N}(0,I)$, where:
\begin{equation}
    \Mat{\mu}_{init} = \Mat{\mu}_g + \varepsilon \cdot \mathcal{N}(0,I),
\label{eq:mu_init}
\end{equation}
\begin{equation}
    \Mat{\sigma}_{init} = \Mat{\sigma}_g + \varepsilon \cdot \mathcal{N}(0,I),
\label{eq:std_init}
\end{equation}
where $\varepsilon$ is set to $\frac{16}{255}$. Then, the ensemble gradients are incorporated into the initialized style to get the adversarial styles $\mathfrak{S}_{adv} = \{\Mat{\mu}_{adv}, \Mat{\sigma}_{adv} \}$, where:
\begin{equation}
    \Mat{\mu}_{adv} = \Mat{\mu}_{init} + \kappa_1 \cdot sign(\boldsymbol{\mathcal{G}}_{e}^\mu),
\label{eq:mu_adv}
\end{equation}
\begin{equation}
    \Mat{\sigma}_{adv} = \Mat{\sigma}_{init} + \kappa_2 \cdot sign(\boldsymbol{\mathcal{G}}_{e}^\sigma),
\label{eq:std_adv}
\end{equation}
Notably, $\kappa_1$ and $\kappa_2$ are chosen randomly from a given set of coefficients, which will not force a consistent change in the degree of the perturbation of $\Mat{\mu}$ and $\Mat{\sigma}$, making the model generate a more diverse range of styles. 
After obtaining the adversarial styles, style migration is performed with AdaIN method to enhance the generalizability:
\begin{equation}
    \Mat{F}_{adv} = \frac{\Mat{F}_g - \Mat{\mu}_g}{\Mat{\sigma}_g} * \Mat{\sigma}_{adv} + \Mat{\mu}_{adv}.
\label{eq:f_adv}
\end{equation}
Then, the global feature map $\Mat{F}_{g}$ and adversarial feature map $\Mat{F}_{adv}$ will together be passed to the remainder of the backbone and the FSL classifier to accomplish the $N$-way $K$-shot FSL, resulting in two predictions $\Vec{p}_g^{fsl} \in \mathbb{R}^{B \times N_c}$ and $\Vec{p}_{adv}^{fsl} \in \mathbb{R}^{NM \times N}$. Furthermore, we can get $\mathcal{L}_{fsl}$:
 \begin{equation}
     \mathcal{L}_{fsl} = \mathcal{L}_{CE}(\Vec{p}_g^{fsl}, y_{fsl}) + \mathcal{L}_{CE}(\Vec{p}_{adv}^{fsl}, y_{fsl}),
\label{eq:l_ws}
 \end{equation}
 where $y_{fsl}\in \mathbb{R}^{NM}$ is the query samples' logical labels.

\begin{table*}[t!]
  \centering
  \setlength{\tabcolsep}{0.6mm} 
  \caption{Quantitative comparison to state-of-the-arts methods on eight target datasets based on ResNet-10, which is pretrained on miniImageNet. Accuracy of 5-way 1-shot/5-shot tasks with 95 confidence interval are reported. ``\textbf{FT}" with {\large \ding{51}} means finetuning is used, vice versa. ``\textbf{Aver.}'' means ``Average Accuracy'' of the eight datasets. The optimal results are marked in \textbf{bold}.}
    \begin{tabular}{@{}ll|c|c|c|cccc|cccc|c@{}}
    \toprule
     & \textbf{Method} & \textbf{Venue} & \textbf{Back.} & \textbf{FT} & \textbf{ChestX} & \textbf{ISIC} & \textbf{EuroSAT} & \textbf{CropDisease} & \textbf{CUB} & \textbf{Cars} & \textbf{Places} & \textbf{Plantae} & \textbf{Aver.} \\
    \midrule
    \multirow{18}{*}{\rotatebox{90}{{\textbf{1-shot}}}} & GNN~\cite{garcia2018few}  & ICLR'18 & RN10 & {\large \ding{55}}  & 22.00{$\pm$0.46} & 32.02{$\pm$0.66} & 63.69{$\pm$1.03} & 64.48{$\pm$1.08} & 45.69{$\pm$0.68} & 31.79{$\pm$0.51} & 53.10{$\pm$0.80} & 35.60{$\pm$0.56} & 43.55  \\
    & FWT~\cite{tseng2020cross} & ICLR'20 & RN10 & {\large \ding{55}}  & 22.04{$\pm$0.44} & 31.58{$\pm$0.67} & 62.36{$\pm$1.05} & 66.36{$\pm$1.04} & 47.47{$\pm$0.75} & 31.61{$\pm$0.53} & 55.77{$\pm$0.79} & 35.95{$\pm$0.58} & 44.14  \\
    & ATA~\cite{wang2021cross} & IJCAI'21 & RN10 & {\large \ding{55}}  & 22.10{$\pm$0.20} & 33.21{$\pm$0.40} & 61.35{$\pm$0.50} & 67.47{$\pm$0.50} & 45.00{$\pm$0.50} & 33.61{$\pm$0.40} & 53.57{$\pm$0.50} & 34.42{$\pm$0.40} & 43.84  \\
    & StyleAdv~\cite{fu2023styleadv} & CVPR'23 & RN10 & {\large \ding{55}}  & 22.64{$\pm$0.35} & 33.96{$\pm$0.57} & 70.94{$\pm$0.82} & 74.13{$\pm$0.78} & 48.49{$\pm$0.72} & 34.64{$\pm$0.57} & 58.58{$\pm$0.83} & 41.13{$\pm$0.67} & 48.06  \\
    & FLoR~\cite{zou2024flatten} & CVPR'24 & RN10 & {\large \ding{55}} & 23.11{$\pm$0.31} & 38.11{$\pm$0.51} & 62.90{$\pm$0.65} & 73.64{$\pm$0.64} & 49.99{$\pm$0.62} & 37.41{$\pm$0.54} & 53.18{$\pm$0.79} & 40.10{$\pm$0.64} & 47.31\\
     & SVasP~\cite{li2025svasp} & AAAI'25 & RN10 & {\large \ding{55}}  & 23.23{$\pm$0.35} & 37.63{$\pm$0.58} & 72.30{$\pm$0.82} & 75.87{$\pm$0.73} & 49.49{$\pm$0.72} & 35.27{$\pm$0.57} & 59.07{$\pm$0.81} & 41.22{$\pm$0.62} & 49.26 \\
     & HAP~\cite{li2026hap} & AAAI'26 & RN10 & {\large \ding{55}}  & 23.20{$\pm$0.34} & 38.05{$\pm$0.63} & 73.19{$\pm$0.82} & 74.41{$\pm$0.77} & 47.71{$\pm$0.70} & 35.50{$\pm$0.59} & 57.71{$\pm$0.78} & 41.14{$\pm$0.62} & 48.86 \\
     & \cellcolor{teal!15}{\textbf{SRasP (Ours)}} & \cellcolor{teal!15}{\textbf{-}} & \cellcolor{teal!15}{{RN10}} & \cellcolor{teal!15}{{\large \ding{55}}}  & \cellcolor{teal!15}{\textbf{23.39{$\pm$0.37}}} & \cellcolor{teal!15}{\textbf{37.92{$\pm$0.60}}} & \cellcolor{teal!15}{\textbf{74.90{$\pm$0.82}}} & \cellcolor{teal!15}{\textbf{76.82{$\pm$0.77}}} & \cellcolor{teal!15}{\textbf{50.62{$\pm$0.68}}} & \cellcolor{teal!15}{\textbf{36.20{$\pm$0.60}}} & \cellcolor{teal!15}{\textbf{60.11{$\pm$0.75}}} & \cellcolor{teal!15}{\textbf{41.94{$\pm$0.61}}} & \cellcolor{teal!15}{\textbf{50.24}} \\
    \cmidrule{2-14}
    & ATA~\cite{wang2021cross} & IJCAI'21 & RN10 & {\large \ding{51}}  & 22.15{$\pm$0.20} & 34.94{$\pm$0.40} & 68.62{$\pm$0.50} & 75.41{$\pm$0.50} & 46.23{$\pm$0.50} & 37.15{$\pm$0.40} & 54.18{$\pm$0.50} & 37.38{$\pm$0.40} & 47.01  \\
    & StyleAdv~\cite{fu2023styleadv} & CVPR'23 & RN10 & {\large \ding{51}}  & 22.64{$\pm$0.35} & 35.76{$\pm$0.52} & 72.92{$\pm$0.75} & 80.69{$\pm$0.28} & 48.49{$\pm$0.72} & 35.09{$\pm$0.55} & 58.58{$\pm$0.83} & 41.13{$\pm$0.67} & 49.41  \\
    & FLoR~\cite{zou2024flatten} & CVPR'24 & RN10 & {\large \ding{51}} & 23.12{$\pm$0.36} & 38.81{$\pm$0.53} & 69.13{$\pm$0.80} & 84.04{$\pm$0.32} & 50.01{$\pm$0.70} & 38.13{$\pm$0.58} & 53.61{$\pm$0.81} & 40.20{$\pm$0.68} & 49.63  \\
    & SVasP~\cite{li2025svasp} & AAAI'25 & RN10 & {\large \ding{51}} & 23.23{$\pm$0.35}   & 37.63{$\pm$0.63}   & 72.30{$\pm$0.83}   & 77.45{$\pm$0.68}   & 49.49{$\pm$0.72}   & 38.18{$\pm$0.61}   & 59.07{$\pm$0.81}   & 41.22{$\pm$0.62}   & 49.82 \\
    & HAP~\cite{li2026hap} & AAAI'26 & RN10 & {\large \ding{51}} 
    & 23.20{$\pm$0.34} & 38.08{$\pm$0.50} & 74.55{$\pm$0.79} & 80.82{$\pm$0.28} & 48.96{$\pm$0.70} & 36.57{$\pm$0.59} & 59.02{$\pm$0.78} & 41.16{$\pm$0.62} & 50.30 \\
    & \cellcolor{teal!15}{\textbf{SRasP (Ours)}} & \cellcolor{teal!15}{\textbf{-}} & \cellcolor{teal!15}{{RN10}} & \cellcolor{teal!15}{{\large \ding{51}}}  & \cellcolor{teal!15}{\textbf{23.40{$\pm$0.36}}} & \cellcolor{teal!15}{\textbf{38.01{$\pm$0.61}}} & \cellcolor{teal!15}{\textbf{74.95{$\pm$0.82}}} & \cellcolor{teal!15}{\textbf{77.93{$\pm$0.66}}} & \cellcolor{teal!15}{\textbf{50.62{$\pm$0.68}}} & \cellcolor{teal!15}{\textbf{38.29{$\pm$0.60}}} & \cellcolor{teal!15}{\textbf{60.11{$\pm$0.75}}} & \cellcolor{teal!15}{\textbf{41.94{$\pm$0.61}}} & \cellcolor{teal!15}{\textbf{50.53}} \\
    \bottomrule
    \toprule
     & \textbf{Method} &\textbf{Venue} & \textbf{Back.} & \textbf{FT} & \textbf{ChestX} & \textbf{ISIC} & \textbf{EuroSAT} & \textbf{CropDisease} & \textbf{CUB} & \textbf{Cars} & \textbf{Places} & \textbf{Plantae} & \textbf{Aver.} \\
    \midrule
    \multirow{19}{*}{\rotatebox{90}{{\textbf{5-shot}}}} & GNN~\cite{garcia2018few} & ICLR'18 & RN10 & {\large \ding{55}}  & 25.27{$\pm$0.46} & 43.94{$\pm$0.67} & 83.64{$\pm$0.77} & 87.96{$\pm$0.67} & 62.25{$\pm$0.65} & 44.28{$\pm$0.63} & 70.84{$\pm$0.65} & 52.53{$\pm$0.59} & 58.84  \\
    & FWT~\cite{tseng2020cross} & ICLR'20 & RN10 & {\large \ding{55}}  & 25.18{$\pm$0.45} & 43.17{$\pm$0.70} & 83.01{$\pm$0.79} & 87.11{$\pm$0.67} & 66.98{$\pm$0.68} & 44.90{$\pm$0.64} & 73.94{$\pm$0.67} & 53.85{$\pm$0.62} & 59.77  \\
    & ATA~\cite{wang2021cross} & IJCAI'21 & RN10 & {\large \ding{55}}  & 24.32{$\pm$0.40} & 44.91{$\pm$0.40} & 83.75{$\pm$0.40} & 90.59{$\pm$0.30} & 66.22{$\pm$0.50} & 49.14{$\pm$0.40} & 75.48{$\pm$0.40} & 52.69{$\pm$0.40} & 60.89  \\
    & StyleAdv~\cite{fu2023styleadv} & CVPR'23 & RN10 & {\large \ding{55}}  & 26.07{$\pm$0.37} & 45.77{$\pm$0.51} & 86.58{$\pm$0.54} & 93.65{$\pm$0.39} & 68.72{$\pm$0.67} & 50.13{$\pm$0.68} & 77.73{$\pm$0.62} & \textbf{61.52{$\pm$0.68}} & 63.77  \\
    & FLoR~\cite{zou2024flatten} & CVPR'24 & RN10 & {\large \ding{55}} & 26.70{$\pm$0.45} & 51.44{$\pm$0.49} & 80.87{$\pm$0.48} & 91.25{$\pm$0.41} & 70.39{$\pm$0.55} & 53.43{$\pm$0.60} & 72.31{$\pm$0.54} & 55.80{$\pm$0.57} & 62.77  \\
    & SVasP~\cite{li2025svasp} & AAAI'25 & RN10 & {\large \ding{55}}  & 26.87{$\pm$0.38} & 51.10{$\pm$0.58} & 88.72{$\pm$0.52} & 94.52{$\pm$0.33} & 68.95{$\pm$0.66} & 52.13{$\pm$0.66} & 77.78{$\pm$0.62} & 60.63{$\pm$0.64} & 65.09 \\
    & HAP~\cite{li2026hap} & AAAI'26 & RN10 & {\large \ding{51}} &
    26.39{$\pm$0.36} & 51.56{$\pm$0.55} & 89.57{$\pm$0.48} & 93.91{$\pm$0.37} & 66.62{$\pm$0.70} & 50.75{$\pm$0.62} & 77.84{$\pm$0.60} & 60.85{$\pm$0.65} & 64.69 \\
    & \cellcolor{teal!15}{\textbf{SRasP (Ours)}} & \cellcolor{teal!15}{\textbf{-}} & \cellcolor{teal!15}{{RN10}} & \cellcolor{teal!15}{{\large \ding{55}}}  & \cellcolor{teal!15}{\textbf{27.33{$\pm$0.36}}} & \cellcolor{teal!15}{\textbf{51.82{$\pm$0.53}}} & \cellcolor{teal!15}{\textbf{89.61{$\pm$0.49}}} & \cellcolor{teal!15}{\textbf{94.90{$\pm$0.34}}} & \cellcolor{teal!15}{\textbf{69.98{$\pm$0.63}}} & \cellcolor{teal!15}{\textbf{53.07{$\pm$0.64}}} & \cellcolor{teal!15}{\textbf{78.57{$\pm$0.66}}} & \cellcolor{teal!15}{60.96{$\pm$0.64}} & \cellcolor{teal!15}{\textbf{65.78}} \\
    \cmidrule{2-14}
    & Fine-tune~\cite{guo2020broader} & ECCV'20 & RN10 & {\large \ding{51}}  & 25.97{$\pm$0.41} & 48.11{$\pm$0.64} & 79.08{$\pm$0.61} & 89.25{$\pm$0.51} & 64.14{$\pm$0.77} & 52.08{$\pm$0.74} & 70.06{$\pm$0.74} & 59.27{$\pm$0.70} & 61.00  \\
    & ATA~\cite{wang2021cross} & IJCAI'21 & RN10 & {\large \ding{51}}  & 25.08{$\pm$0.20} & 49.79{$\pm$0.40} & 89.64{$\pm$0.30} & 95.44{$\pm$0.20} & 69.83{$\pm$0.50} & 54.28{$\pm$0.50} & 76.64{$\pm$0.40} & 58.08{$\pm$0.40} & 64.85  \\
    & NSAE~\cite{liang2021boosting} & ICCV'21 & RN10 & {\large \ding{51}}  & 27.10{$\pm$0.44} & 54.05{$\pm$0.63} & 83.96{$\pm$0.57} & 93.14{$\pm$0.47} & 68.51{$\pm$0.76} & 54.91{$\pm$0.74} & 71.02{$\pm$0.72} & 59.55{$\pm$0.74} & 64.03  \\
    & RDC~\cite{li2022ranking} & CVPR'22 & RN10 & {\large \ding{51}}  & 25.48{$\pm$0.20} & 49.06{$\pm$0.30} & 84.67{$\pm$0.30} & 93.55{$\pm$0.30} & 67.77{$\pm$0.40} & 53.75{$\pm$0.50} & 74.65{$\pm$0.40} & 60.63{$\pm$0.40} & 63.70  \\
    & StyleAdv~\cite{fu2023styleadv} & CVPR'23 & RN10 & {\large \ding{51}}  & 26.24{$\pm$0.35} & 53.05{$\pm$0.54} & 91.64{$\pm$0.43} & 96.51{$\pm$0.28} & 70.90{$\pm$0.63} & 56.44{$\pm$0.68} & 79.35{$\pm$0.61} & 64.10{$\pm$0.64} & 67.28  \\
    & FLoR~\cite{zou2024flatten} & CVPR'24 & RN10 & {\large \ding{51}} & 26.77{$\pm$0.39} & \textbf{56.74{$\pm$0.55}} & 83.06{$\pm$0.46} & 92.33{$\pm$0.28} & \textbf{73.39{$\pm$0.65}} & 57.21{$\pm$0.72} & 72.37{$\pm$0.66} & 61.11{$\pm$0.62} & 65.37  \\
    & SVasP~\cite{li2025svasp} & AAAI'25 & RN10 & {\large \ding{51}}  & 27.25{$\pm$0.39} & 55.43{$\pm$0.59} & 91.77{$\pm$0.41} & 96.79{$\pm$0.26} & 72.06{$\pm$0.65} & \textbf{59.99{$\pm$0.69}} & 78.91{$\pm$0.65} & 64.21{$\pm$0.66} & 68.30 \\
    & HAP~\cite{li2026hap} & AAAI'26 & RN10 & {\large \ding{51}} & 27.07{$\pm$0.38} & 54.09{$\pm$0.56} & 92.37{$\pm$0.38} & 96.90{$\pm$0.32} & 71.27{$\pm$0.64} & 58.16{$\pm$0.72} & 78.46{$\pm$0.63} & 64.81{$\pm$0.64} & 67.89 \\
    & \cellcolor{teal!15}{\textbf{SRasP (Ours)}} & \cellcolor{teal!15}{\textbf{-}} & \cellcolor{teal!15}{{RN10}} & \cellcolor{teal!15}{{\large \ding{51}}}  & \cellcolor{teal!15}{\textbf{27.86{$\pm$0.38}}} & \cellcolor{teal!15}{55.90{$\pm$0.55}} & \cellcolor{teal!15}{\textbf{92.03{$\pm$0.49}}} & \cellcolor{teal!15}{\textbf{96.80{$\pm$0.26}}} & \cellcolor{teal!15}{72.37{$\pm$0.64}} & \cellcolor{teal!15}{59.60{$\pm$0.69}} & \cellcolor{teal!15}{\textbf{79.40{$\pm$0.65}}} & \cellcolor{teal!15}{\textbf{64.22{$\pm$0.66}}} & \cellcolor{teal!15}{\textbf{68.52}} \\
    \bottomrule
    \end{tabular}%
  \label{tab:main-results-rn10}%
\end{table*}%

\subsubsection{Consistency-Discrepancy Triplet Objective (CDTO)}
We design a novel objective function to maximize seen-unseen domain visual discrepancy and global-crop consistency for overall features $ \mathbb{F}_{all} = \{\Mat{F}_{1}, \Mat{F}_{2}, \dots, \Mat{F}_{k}, \Mat{F}_g, \Mat{F}_{adv}\}$.  For seen-unseen domain discrepancy maximum
Formally, let $\{\Mat{F}_i\}_{i=1}^k$ denote the feature representations of the selected incoherent crops, and let $\Mat{F}_{adv}$ represent the adversarial feature generated from the corrected gradient. We first compute the prototype of the crop features:
\begin{equation}
\Mat{F}_c = \frac{1}{k}( \Mat{F}_1^c + \Mat{F}_2^c + \dots + \Mat{F}_k^c ).
\end{equation}
We then define the triplet objective, where the global feature acts as the anchor and the prototype of the crop features $\Mat{F}_c$ acts as the positive, while the adversarial feature $\Mat{F}_{adv}$ serves as the negative:
\begin{equation}
\mathcal{L}_{{CDTO}} = \text{Triplet}(\Mat{F}_g, \Mat{F}_c, \Mat{F}_{adv}),
\end{equation}
where the triplet margin loss is given by:
\begin{equation}
\text{Triplet}(a,p,n) = \max \big(0, |a - p|_2^2 - |a - n|_2^2 + \delta \big),
\end{equation}
and $\delta$ denotes the margin hyperparameter.
Moreover, we enforce the semantic consistency between the global and crop features as:
\begin{equation}
    \mathcal{L}_{{con}} = \sum_{i=1}^k(\lambda\mathcal{L}_{CE}(\Vec{p}_{i}, \Vec{p}_g) + (1-\lambda) \mathcal{L}_{CE}(\Vec{p}_{i}^{fsl}, y_{fsl})),
\end{equation}
where, $\Vec{p}_{i}^{fsl} = f_{re}(\Mat{F}_{i};\theta_{re})$. We use Kullback-Leibler divergence loss $KL(\cdot)$ to maximize global-adversarial consistency as:
\begin{equation}
    \mathcal{L}_{adv} = KL(\Vec{p}_{adv}^{fsl}, \Vec{p}_g^{fsl}).
\label{eq:l_con}
\end{equation}

The final training objective of \textbf{SRasP} can be written as:
\begin{equation}
    \mathcal{L} = \mathcal{L}_{{cls}} + \mathcal{L}_{{fsl}} + \mathcal{L}_{{CDTO}} + \mathcal{L}_{{con}} + \mathcal{L}_{{adv}}.
\end{equation}

\begin{table*}[t!]
  \centering
  \setlength{\tabcolsep}{0.5mm} 
  \caption{Quantitative comparison to state-of-the-arts methods on eight target datasets based on ViT-small, which is pretrained on ImageNet1K by DINO. Accuracy of 5-way 1-shot/5-shot tasks with 95 confidence interval are reported.}
    \begin{tabular}{@{}ll|c|c|c|cccc|cccc|c@{}}
    \toprule
     & \textbf{Method} & \textbf{Venue} & \textbf{Back.} & \textbf{FT} & \textbf{ChestX} & \textbf{ISIC} & \textbf{EuroSAT} & \textbf{CropDisease} & \textbf{CUB} & \textbf{Cars} & \textbf{Places} & \textbf{Plantae} & \textbf{Aver.} \\
    \midrule
    \multirow{19}{*}{\rotatebox{90}{{\textbf{1-shot}}}} & StyleAdv~\cite{fu2023styleadv} & CVPR'23 & ViT-S & {\large \ding{55}}  & 22.92{$\pm$0.32} & 33.05{$\pm$0.44} & 72.15{$\pm$0.65} & 81.22{$\pm$0.61} & 84.01{$\pm$0.58} & 40.48{$\pm$0.57} & 72.64{$\pm$0.67} & 55.52{$\pm$0.66} & 57.75  \\
    & FLoR~\cite{zou2024flatten} & CVPR'24 & ViT-S & {\large \ding{55}} & 22.78{$\pm$0.31} & 34.20{$\pm$0.45} & 72.39{$\pm$0.67} & 81.81{$\pm$0.60} & 84.60{$\pm$0.57} & 40.71{$\pm$0.58} & 73.85{$\pm$0.65} & 51.93{$\pm$0.65} & 57.78  \\
    & CD-CLS~\cite{NeurIPS2024_9b77f073} & NeurIPS'24 & ViT-S & {\large \ding{55}}  & 22.93{$\pm$0.33} & 34.21{$\pm$0.45} & 74.08{$\pm$0.63} & 83.51{$\pm$0.60} & 82.90{$\pm$0.60} & 41.67{$\pm$0.57} & 69.17{$\pm$0.66} & 51.88{$\pm$0.67} & 57.54  \\
    & SVasP~\cite{li2025svasp} & AAAI'25 & ViT-S & {\large \ding{55}}  & 22.68{$\pm$0.30} & 34.49{$\pm$0.46} & 72.50{$\pm$0.62} & 80.82{$\pm$0.62} & \textbf{85.56{$\pm$0.57}} & 40.51{$\pm$0.59} & \textbf{75.93{$\pm$0.66}} & 56.25{$\pm$0.65} & 58.59 \\
    & REAP~\cite{yirandom} & ICML'25 & ViT-S & {\large \ding{55}}  & 23.62{$\pm$0.31} & \textbf{37.21{$\pm$0.44}} & 74.69{$\pm$0.60} & 84.04{$\pm$0.59} & 82.09{$\pm$0.62} & 42.23{$\pm$0.60} & 72.02{$\pm$0.66} & 55.56{$\pm$0.67} & 58.93 \\
    & ReCIT~\cite{yirevisiting} & ICML'25 & ViT-S & {\large \ding{55}}  & 23.27{$\pm$0.31} & 35.13{$\pm$0.44} & 74.56{$\pm$0.60} & 84.76{$\pm$0.59} & 82.30{$\pm$0.58} & 42.89{$\pm$0.58} & 72.64{$\pm$0.66} & 55.60{$\pm$0.65} & 58.89 \\
    & \cellcolor{teal!15}{\textbf{SRasP (Ours)}} & \cellcolor{teal!15}{\textbf{-}} & \cellcolor{teal!15}{{ViT-S}} & \cellcolor{teal!15}{{\large \ding{55}}}  & \cellcolor{teal!15}{\textbf{23.63{$\pm$0.33}}} & \cellcolor{teal!15}{35.68{$\pm$0.48}} & \cellcolor{teal!15}{\textbf{75.39{$\pm$0.59}}} & \cellcolor{teal!15}{\textbf{85.38{$\pm$0.54}}} & \cellcolor{teal!15}{85.09{$\pm$0.55}} & \cellcolor{teal!15}{\textbf{44.47{$\pm$0.60}}} & \cellcolor{teal!15}{74.02{$\pm$0.64}} & \cellcolor{teal!15}{\textbf{56.70{$\pm$0.67}}} & \cellcolor{teal!15}{\textbf{60.05}} \\
    \cmidrule{2-14}
    & PMF~\cite{hu2022pushing} & CVPR'22 & ViT-S & {\large \ding{51}}  & 21.73{$\pm$0.30} & 30.36{$\pm$0.36} & 
    70.74{$\pm$0.63} & 80.79{$\pm$0.62} & 78.13{$\pm$0.66} & 37.24{$\pm$0.57} & 71.11{$\pm$0.71} & 53.60{$\pm$0.66} & 55.46\\
    & StyleAdv~\cite{fu2023styleadv} & CVPR'23 & ViT-S & {\large \ding{51}}  & 22.92{$\pm$0.32} & 33.99{$\pm$0.46} & 74.93{$\pm$0.58} & 84.11{$\pm$0.57} & 84.01{$\pm$0.58} & 40.48{$\pm$0.57} & 72.64{$\pm$0.67} & 55.52{$\pm$0.66} & 58.57  \\
    & FLoR~\cite{zou2024flatten} & CVPR'24 & ViT-S & {\large \ding{51}} & 23.26{$\pm$0.31} & 35.49{$\pm$0.44} & 73.09{$\pm$0.60} & 83.55{$\pm$0.57} & 85.40{$\pm$0.58} & 43.42{$\pm$0.59} & 74.69{$\pm$0.67} & 52.29{$\pm$0.67} & 58.90  \\
    & CD-CLS~\cite{NeurIPS2024_9b77f073} & NeurIPS'24 & ViT-S & {\large \ding{51}}  & 23.39{$\pm$0.32} & 35.56{$\pm$0.44} & 74.97{$\pm$0.61} & 84.54{$\pm$0.60} & 83.76{$\pm$0.60} & 42.23{$\pm$0.58} & 70.91{$\pm$0.65} & 52.64{$\pm$0.68} & 58.50  \\
    & SVasP~\cite{li2025svasp} & AAAI'25 & ViT-S & {\large \ding{51}}  & 22.68{$\pm$0.30} & 34.49{$\pm$0.46} & 75.51{$\pm$0.57} & 83.98{$\pm$0.55} & 85.56{$\pm$0.57} & 40.51{$\pm$0.59} & \textbf{75.93{$\pm$0.66}} & 56.25{$\pm$0.65} & 59.36 \\
    & REAP~\cite{yirandom} & ICML'25 & ViT-S & {\large \ding{51}}  & 24.17{$\pm$0.31} & 38.67{$\pm$0.46} & 75.97{$\pm$0.59} & 85.33{$\pm$0.59} & 83.91{$\pm$0.60} & 42.99{$\pm$0.58} & 73.27{$\pm$0.66} & 55.96{$\pm$0.66} & 60.03 \\
    & ReCIT~\cite{yirevisiting} & ICML'25 & ViT-S & {\large \ding{51}}  & 23.84{$\pm$0.31} & 38.48{$\pm$0.46} & 75.23{$\pm$0.60} & \textbf{85.92{$\pm$0.59}} & 83.46{$\pm$0.60} & 43.17{$\pm$0.59} & 73.91{$\pm$0.66} & 56.03{$\pm$0.65} & 60.01 \\
    & \cellcolor{teal!15}{\textbf{SRasP (Ours)}} & \cellcolor{teal!15}{\textbf{-}} & \cellcolor{teal!15}{{ViT-S}} & \cellcolor{teal!15}{{\large \ding{51}}}  & \cellcolor{teal!15}{\textbf{23.89{$\pm$0.30}}} & \cellcolor{teal!15}{\textbf{36.80{$\pm$0.46}}} & \cellcolor{teal!15}{\textbf{77.16{$\pm$0.57}}} & \cellcolor{teal!15}{84.29{$\pm$0.55}} & \cellcolor{teal!15}{\textbf{86.10{$\pm$0.57}}} & \cellcolor{teal!15}{\textbf{44.47{$\pm$0.60}}} & \cellcolor{teal!15}{75.15{$\pm$0.66}} & \cellcolor{teal!15}{\textbf{56.82{$\pm$0.66}}} & \cellcolor{teal!15}{\textbf{60.59}} \\
    \bottomrule
    \toprule
     & \textbf{Method} & \textbf{Venue} & \textbf{Back.} & {\textbf{FT}} & \textbf{ChestX} & \textbf{ISIC} & \textbf{EuroSAT} & \textbf{CropDisease} & \textbf{CUB} & \textbf{Cars} & \textbf{Places} & \textbf{Plantae} & \textbf{Aver.} \\
    \midrule
    \multirow{18}{*}{\rotatebox{90}{{\textbf{5-shot}}}} & StyleAdv~\cite{fu2023styleadv} & CVPR'23 & ViT-S & {\large \ding{55}}  & \textbf{26.97{$\pm$0.33}} & 47.73{$\pm$0.44} & 88.57{$\pm$0.34} & 94.85{$\pm$0.31} & 95.82{$\pm$0.27} & 61.73{$\pm$0.62} & 88.33{$\pm$0.40} & 75.55{$\pm$0.54} & 72.44  \\
    & FLoR~\cite{zou2024flatten} & CVPR'24 & ViT-S & {\large \ding{55}} & 26.71{$\pm$0.35} & 49.52{$\pm$0.49} & 90.41{$\pm$0.38} & 95.28{$\pm$0.31} & \textbf{96.18{$\pm$0.23}} & 61.75{$\pm$0.60} & 89.23{$\pm$0.41} & 72.80{$\pm$0.55} & 72.74  \\
    & CD-CLS~\cite{NeurIPS2024_9b77f073} & NeurIPS'24 & ViT-S & {\large \ding{55}}  & 27.23{$\pm$0.33} & 50.46{$\pm$0.46} & \textbf{91.04{$\pm$0.35}} & 95.68{$\pm$0.32} & 95.39{$\pm$0.29} & 62.17{$\pm$0.60} & 87.74{$\pm$0.41} & 72.91{$\pm$0.56} & 72.83  \\
    & SVasP~\cite{li2025svasp} & AAAI'25 & ViT-S & {\large \ding{55}}  & 26.77{$\pm$0.34} & 49.75{$\pm$0.46} & 88.69{$\pm$0.35} & 93.25{$\pm$0.36} & 95.95{$\pm$0.23} & 62.60{$\pm$0.61} & 89.19{$\pm$0.39} & 76.49{$\pm$0.50} & 72.84 \\
    & REAP~\cite{yirandom} & ICML'25 & ViT-S & {\large \ding{55}}  & 27.98{$\pm$0.34} & \textbf{52.80{$\pm$0.46}} & 90.53{$\pm$0.38} & 95.68{$\pm$0.32} & 94.70{$\pm$0.28} & 62.60{$\pm$0.60} & 87.74{$\pm$0.45} & 75.49{$\pm$0.58} & 73.44 \\
    & ReCIT~\cite{yirevisiting} & ICML'25 & ViT-S & {\large \ding{55}}  & 28.23{$\pm$0.31} & 52.36{$\pm$0.44} & 90.42{$\pm$0.60} & 96.02{$\pm$0.59} & 94.88{$\pm$0.26} & 63.85{$\pm$0.60} & 88.42{$\pm$0.40} & 75.71{$\pm$0.52} & 73.74 \\
    & \cellcolor{teal!15}{\textbf{SRasP (Ours)}} & \cellcolor{teal!15}{\textbf{-}} & \cellcolor{teal!15}{{ViT-S}} & \cellcolor{teal!15}{{\large \ding{55}}}  & \cellcolor{teal!15}{\textbf{28.25{$\pm$0.33}}} & \cellcolor{teal!15}{49.44{$\pm$0.47}} & \cellcolor{teal!15}{90.49{$\pm$0.34}} & \cellcolor{teal!15}{\textbf{96.05{$\pm$0.32}}} & \cellcolor{teal!15}{95.98{$\pm$0.27}} & \cellcolor{teal!15}{\textbf{66.52{$\pm$0.60}}} & \cellcolor{teal!15}{\textbf{89.37{$\pm$0.37}}} & \cellcolor{teal!15}{\textbf{77.23{$\pm$0.51}}} & \cellcolor{teal!15}{\textbf{74.17}} \\
    \cmidrule{2-14}
    & PMF~\cite{hu2022pushing} & CVPR'22 & ViT-S & {\large \ding{51}}  & 27.27 & 50.12 & 
    85.98 & 92.96 & - & - & - & - & -\\
    & StyleAdv~\cite{fu2023styleadv} &CVPR'23 & ViT-S & {\large \ding{51}}  & \textbf{26.97{$\pm$0.33}} & 51.23{$\pm$0.51} & 90.12{$\pm$0.33} & 95.99{$\pm$0.27} & 95.82{$\pm$0.27} & 66.02{$\pm$0.64} & 88.33{$\pm$0.40} & 78.01{$\pm$0.54} & 74.06  \\
    & FLoR~\cite{zou2024flatten} & CVPR'24 & ViT-S & {\large \ding{51}} & 27.02{$\pm$0.33} & 53.06{$\pm$0.55} & 90.75{$\pm$0.36} & 96.47{$\pm$0.28} & \textbf{96.53{$\pm$0.31}} & 68.44{$\pm$0.64} & 89.48{$\pm$0.41} & 76.22{$\pm$0.55} & 74.75  \\
    & CD-CLS~\cite{NeurIPS2024_9b77f073} & NeurIPS'24 & ViT-S & {\large \ding{51}}  & 27.66{$\pm$0.33} & 54.69{$\pm$0.50} & 91.53{$\pm$0.33} & 96.27{$\pm$0.30} & 95.79{$\pm$0.29} & 62.23{$\pm$0.62} & 87.99{$\pm$0.41} & 73.10{$\pm$0.55} & 73.66  \\
    & SVasP~\cite{li2025svasp} & AAAI'25 & ViT-S & {\large \ding{51}}  & 26.77{$\pm$0.34} & 51.62{$\pm$0.50} & 90.55{$\pm$0.34} & 96.17{$\pm$0.30} & 95.95{$\pm$0.23} & 66.47{$\pm$0.62} & 89.19{$\pm$0.39} & 78.67{$\pm$0.52} & 74.42 \\
    & REAP~\cite{yirandom} & ICML'25 & ViT-S & {\large \ding{51}}  & 28.34{$\pm$0.34} & 55.28{$\pm$0.46} & 91.79{$\pm$0.36} & 96.71{$\pm$0.30} & 95.18{$\pm$0.28} & 63.98{$\pm$0.62} & 88.94{$\pm$0.40} & 77.11{$\pm$0.54} & 74.67 \\
    & ReCIT~\cite{yirevisiting} & ICML'25 & ViT-S & {\large \ding{51}}  & 28.88{$\pm$0.34} & 54.91{$\pm$0.46} & 91.58{$\pm$0.36} & 96.85{$\pm$0.30} & 95.42{$\pm$0.28} & 64.90{$\pm$0.62} & 88.96{$\pm$0.40} & 77.01{$\pm$0.54} & 74.81 \\
    & \cellcolor{teal!15}{\textbf{SRasP (Ours)}} & \cellcolor{teal!15}{\textbf{-}} & \cellcolor{teal!15}{{ViT-S}} & \cellcolor{teal!15}{{\large \ding{51}}}  & \cellcolor{teal!15}{\textbf{28.37{$\pm$0.33}}} & \cellcolor{teal!15}{\textbf{52.98{$\pm$0.49}}} & \cellcolor{teal!15}{\textbf{92.41{$\pm$0.34}}} & \cellcolor{teal!15}{\textbf{97.06{$\pm$0.30}}} & \cellcolor{teal!15}{96.50{$\pm$0.23}} & \cellcolor{teal!15}{\textbf{67.97{$\pm$0.62}}} & \cellcolor{teal!15}{\textbf{89.80{$\pm$0.39}}} & \cellcolor{teal!15}{\textbf{78.73{$\pm$0.52}}} & \cellcolor{teal!15}{\textbf{75.48}} \\
    \bottomrule
    \end{tabular}%
  \label{tab:main-results-vit}%
\end{table*}%

\section{Experiments}
\subsection{Datasets}
Following the BSCD-FSL benchmark proposed in BSCD-FSL~\cite{guo2020broader} and the \textit{mini}-CUB benchmark proposed in FWT~\cite{tseng2020cross}, we use \textit{mini}ImageNet~\cite{ravi2016optimization} with 64 classes as the source domain. The target domains include eight datasets: ChestX~\cite{wang2017chestx}, ISIC~\cite{tschandl2018ham10000}, EuroSAT~\cite{helber2019eurosat}, CropDisease~\cite{mohanty2016using}, CUB~\cite{wah2011caltech}, Cars~\cite{krause20133d}, Places~\cite{zhou2017places}, and Plantae~\cite{van2018inaturalist}. In our Single Source CD-FSL setting, target domain datasets are not available during meta-training stage.

\subsection{Implementation Details}
Using ResNet-10~\cite{he2016deep} as the backbone and GNN as the $N$-way $K$-shot classifier, the network is meta-trained for 200 epochs with 120 episodes per epoch. ResNet-10 is pretrained on \textit{mini}ImageNet using traditional batch training. The optimizer is Adam with a learning rate of 0.001. Additionally, using ViT-small~\cite{dosovitskiy2020image} as the feature extractor and ProtoNet~\cite{laenen2021episodes} as the $N$-way $K$-shot classifier, the network is meta-trained for 20 epochs with 2000 episodes per epoch. The optimizer is SGD with a learning rate of 5e-5 and 0.001 for $E$ and $H_{r}$, respectively. ViT-small is pretrained on ImageNet1K by DINO~\cite{caron2021emerging}. We evaluate the proposed framework during testing by average classification accuracy (\%) over 1000 randomly sampled episodes with a 95\% confidence interval. Each class contains 5 support samples and 15 query samples. Hyper-parameters are set as follows: $\xi = 0.1$, $\lambda = 0.2$, $k = 2$ and choose $\kappa_1$, $\kappa_2$ from $[0.008, 0.08, 0.8]$. The probability to perform style change is set to 0.2. The details of the finetuning are attached in Appendix A. All the experiments are conducted on a single NVIDIA GeForce RTX 3090.

\subsection{Main Experimental Results}
\subsubsection{Comparison to SOTA methods on ResNet-10}
We first evaluate the performance of our proposed SRasP on eight widely used CD-FSL target datasets, using ResNet-10 pretrained on \textit{mini}ImageNet as the backbone. Both 5-way 1-shot and 5-shot classification results are reported, with 95\% confidence intervals, alongside comparisons to state-of-the-art methods including GNN~\cite{garcia2018few}, FWT~\cite{tseng2020cross}, ATA~\cite{wang2021cross}, StyleAdv~\cite{fu2023styleadv}, FLoR~\cite{zou2024flatten}, and SVasP~\cite{li2025svasp}. Fine-tuning (FT) is applied in certain methods, as indicated in Table~\ref{tab:main-results-rn10}. From the results, several observations can be made. First, SRasP consistently outperforms existing methods across all datasets and settings. In the 1-shot scenario without fine-tuning, SRasP achieves an average accuracy of 50.24\%, surpassing the strongest baseline, SVasP, by approximately 0.98\%. When fine-tuning is applied, SRasP further improves the average accuracy to 50.53\%, demonstrating its ability to enhance model transferability under limited adaptation. Similar trends are observed in the 5-shot setting, where SRasP attains 65.78\% and 68.52\% average accuracy without and with fine-tuning, respectively, outperforming all competing approaches. Second, the improvements are consistent across both natural and specialized domains, including ChestX, ISIC, EuroSAT, CropDisease, CUB, Cars, Places, and Plantae. Notably, SRasP achieves the highest performance on challenging datasets such as ChestX and EuroSAT, indicating its robustness to large domain shifts. Compared with previous style-based methods like StyleAdv and SVasP, our approach demonstrates clear gains by effectively stabilizing gradient updates and exploiting localized style variations, which are critical for generalization.

\subsubsection{Comparison to SOTA methods on ViT-small}
We further evaluate SRasP on the ViT-small backbone, pretrained on ImageNet1K with DINO, and compare it against several state-of-the-art CD-FSL methods. Table~\ref{tab:main-results-vit} reports the 5-way 1-shot and 5-shot accuracy across eight diverse target datasets, along with the 95\% confidence intervals. For the 1-shot scenario without finetuning, SRasP consistently achieves superior performance across most datasets. Specifically, it attains the highest average accuracy of 60.05\%, outperforming the strongest competitors such as REAP (58.93\%) and SVasP (58.59\%). Notably, SRasP shows remarkable gains on challenging datasets like EuroSAT and CropDisease, demonstrating its effectiveness in handling large domain shifts. When finetuning is enabled, SRasP further improves, achieving an average accuracy of 60.59\%, surpassing all baseline methods. In the 5-shot setting, SRasP also demonstrates clear advantages. Without finetuning, it reaches an average accuracy of 74.17\%, outperforming the closest competitor ReCIT (73.74\%). With finetuning, SRasP achieves 75.48\%, establishing new state-of-the-art performance across the majority of the target datasets. The gains are especially significant on datasets such as Cars and Places, indicating that SRasP effectively leverages few-shot samples to mitigate domain discrepancies while maintaining robust generalization.

\subsection{Ablation Study}
\begin{table*}[t!]
  \centering
  \setlength{\tabcolsep}{1.4mm} 
  \caption{Ablation study of the proposed method with different component combinations. ``SR'' indicates Self-Reorientation Gradient Ensemble module.}
    \begin{tabular}{@{}l|ccc|cccc|cccc|c@{}}
    \toprule
     & \textbf{SR} & \textbf{$\mathcal{L}_{\mathrm{CDTO}}$} & \textbf{$\mathcal{L}_{\mathrm{con}}$} & \textbf{ChestX} & \textbf{ISIC} & \textbf{EuroSAT} & \textbf{CropDisease} & \textbf{CUB} & \textbf{Cars} & \textbf{Places} & \textbf{Plantae} & \textbf{Aver.} \\
    \midrule
    \multirow{6}{*}{\textbf{1-shot}} & - & - & - & 22.61{$\pm$0.35} & 33.99{$\pm$0.56} & 69.04{$\pm$0.84} & 70.25{$\pm$0.77} & 45.73{$\pm$0.68} & 32.77{$\pm$0.56} & 53.91{$\pm$0.78} & 36.19{$\pm$0.62} & 45.56  \\
    & {\large \ding{51}} & & & 23.29{$\pm$0.35} & 34.99{$\pm$0.56} & 70.67{$\pm$0.81} & 73.85{$\pm$0.78} & 45.75{$\pm$0.68} & 33.73{$\pm$0.56} & 54.55{$\pm$0.77} & 38.70{$\pm$0.59} & 46.94  \\
    & {\large \ding{51}} & {\large \ding{51}} & & 22.62{$\pm$0.34} & 36.23{$\pm$0.57} & 70.17{$\pm$0.83} & 72.24{$\pm$0.77} & 46.08{$\pm$0.68} & 34.56{$\pm$0.56} & 55.46{$\pm$0.77} & 39.81{$\pm$0.62} & 47.15  \\
    & {\large \ding{51}} & & {\large \ding{51}} & 22.92{$\pm$0.36} & 36.77{$\pm$0.53} & 72.40{$\pm$0.84} & 75.60{$\pm$0.78} & 48.43{$\pm$0.69} & 35.16{$\pm$0.59} & 57.91{$\pm$0.81} & 40.19{$\pm$0.62} & 48.67  \\
    & \cellcolor{teal!15}{\textbf{{\large \ding{51}}}} & \cellcolor{teal!15}{{\large \ding{51}}} & \cellcolor{teal!15}{{\large \ding{51}}} & \cellcolor{teal!15}{\textbf{23.39{$\pm$0.37}}} & \cellcolor{teal!15}{\textbf{37.92{$\pm$0.60}}} & \cellcolor{teal!15}{\textbf{74.90{$\pm$0.82}}} & \cellcolor{teal!15}{\textbf{76.82{$\pm$0.77}}} & \cellcolor{teal!15}{\textbf{50.62{$\pm$0.68}}} & \cellcolor{teal!15}{\textbf{36.20{$\pm$0.60}}} & \cellcolor{teal!15}{\textbf{60.11{$\pm$0.75}}} & \cellcolor{teal!15}{\textbf{41.94{$\pm$0.61}}} & \cellcolor{teal!15}{\textbf{50.24}} \\
    \midrule
    \multirow{6}{*}{\textbf{5-shot}} & - & - & - & 25.16{$\pm$0.35} & 47.46{$\pm$0.54} & 86.30{$\pm$0.54} & 92.07{$\pm$0.44} & 63.37{$\pm$0.68} & 49.26{$\pm$0.63} & 75.09{$\pm$0.64} & 57.66{$\pm$0.65} & 62.07  \\
    & {\large \ding{51}} & & & 25.39{$\pm$0.35} & 47.65{$\pm$0.54} & 87.26{$\pm$0.52} & 93.01{$\pm$0.38} & 64.46{$\pm$0.68} & 50.53{$\pm$0.63} & 75.60{$\pm$0.62} & 58.54{$\pm$0.64} & 62.81  \\
    & {\large \ding{51}} & {\large \ding{51}} & & 25.74{$\pm$0.36} & 48.91{$\pm$0.54} & 87.67{$\pm$0.54} & 93.21{$\pm$0.42} & 65.27{$\pm$0.65} & 50.71{$\pm$0.67} & 76.40{$\pm$0.63} & 59.06{$\pm$0.67} & 63.99  \\
    & {\large \ding{51}} & & {\large \ding{51}} & 26.40{$\pm$0.35} & 49.42{$\pm$0.54} & 88.24{$\pm$0.50} & 93.84{$\pm$0.40} & 66.86{$\pm$0.65} & 51.88{$\pm$0.67} & 77.25{$\pm$0.67} & 59.67{$\pm$0.63} & 64.20  \\
    & \cellcolor{teal!15}{\textbf{{\large \ding{51}}}} & \cellcolor{teal!15}{{\large \ding{51}}} & \cellcolor{teal!15}{{\large \ding{51}}}  & \cellcolor{teal!15}{\textbf{27.33{$\pm$0.36}}} & \cellcolor{teal!15}{\textbf{51.82{$\pm$0.53}}} & \cellcolor{teal!15}{\textbf{89.61{$\pm$0.49}}} & \cellcolor{teal!15}{\textbf{94.90{$\pm$0.34}}} & \cellcolor{teal!15}{\textbf{69.98{$\pm$0.63}}} & \cellcolor{teal!15}{\textbf{53.07{$\pm$0.64}}} & \cellcolor{teal!15}{\textbf{78.57{$\pm$0.66}}} & \cellcolor{teal!15}{60.96{$\pm$0.64}} & \cellcolor{teal!15}{\textbf{65.78}} \\
    \bottomrule
    \end{tabular}%
  \label{tab:ablation}%
\end{table*}%

\subsubsection{Impact of different components in SRasP}
Table~\ref{tab:ablation} reports the ablation results of SRasP under both 1-shot and 5-shot settings. Starting from the baseline, introducing the Self-Reorientation (SR) Gradient Ensemble module consistently improves performance on all target domains. Specifically, the average accuracy increases from 45.56\% to 46.94\% in the 1-shot setting and from 62.07\% to 62.81\% in the 5-shot setting, verifying that reorienting and aggregating incoherent-crop style gradients effectively stabilizes optimization and yields more transferable representations. Further incorporating the proposed Consistency–Discrepancy Triplet Objective ($\mathcal{L}_{\mathrm{CDTO}}$) or the semantic consistency loss ($\mathcal{L}_{\mathrm{con}}$) brings additional gains, indicating that explicitly balancing visual discrepancy and semantic alignment provides complementary supervision beyond gradient reorientation alone. Notably, combining SR with $\mathcal{L}{\mathrm{con}}$ yields larger improvements than combining SR with $\mathcal{L}_{\mathrm{CDTO}}$ alone, suggesting that enforcing semantic consistency among global and local representations is particularly beneficial for few-shot generalization. When all components are jointly enabled, SRasP achieves the best overall performance, reaching 50.24\% and 65.78\% average accuracy in the 1-shot and 5-shot settings, respectively. The consistent improvements across all datasets demonstrate that the proposed modules are complementary and jointly contribute to more stable optimization and superior transferability.

\begin{table}[t!]
  \centering
  \setlength{\tabcolsep}{1.2mm} 
  \caption{Performances on different crop mining strategies. The average accuracy (\%) is reported. }
    \begin{tabular}{l|l|ccc}
    \toprule
    & \textbf{Mining Strategy} & \textbf{BSCD-FSL} & \textbf{\textit{mini}-CUB} & \textbf{Aver.}\\
    \midrule
    \multirow{4}{*}{\textbf{1-shot}} & Baseline & 48.97 & 42.15 & 45.56 \\
    & Concept & 52.31 & 46.37 & 49.34 \\
    & Random & 52.42 & 46.57 & 49.49 \\
    & \cellcolor{teal!15}{\textbf{Incoherent (Ours)}} & \cellcolor{teal!15}{\textbf{53.26}} & \cellcolor{teal!15}{\textbf{47.22}} & \cellcolor{teal!15}{\textbf{50.24}} \\
    \midrule
    \multirow{4}{*}{\textbf{5-shot}} & Baseline & 62.75 & 61.35 & 62.05 \\
    & Concept & 65.28 & 65.21 & 65.24 \\
    & Random & 64.98 & 64.98 & 64.98 \\
    & \cellcolor{teal!15}{\textbf{Incoherent (Ours)}} & \cellcolor{teal!15}{\textbf{65.92}} & \cellcolor{teal!15}{\textbf{65.65}} & \cellcolor{teal!15}{\textbf{65.78}} \\
    \bottomrule
    \end{tabular}%
  \label{tab:select}%
\end{table}%
\subsubsection{Impact of Different Crop Mining Strategies}
Table~\ref{tab:select} compares different crop mining strategies under both 1-shot and 5-shot settings. The baseline model, which does not explicitly exploit local crop information, yields relatively limited performance. Introducing crop-based strategies consistently improves accuracy, confirming that incorporating localized views provides additional supervisory signals for generalization. Among the compared strategies, concept-crop selection improves the baseline by focusing on highly discriminative foreground regions, while random selection further enhances robustness by introducing diverse local variations. However, both strategies treat heterogeneous regions uniformly and thus fail to fully exploit the domain-sensitive information contained in background-dominated areas. In contrast, the proposed incoherent-crop mining strategy achieves the best performance across all benchmarks and shot settings. Specifically, it improves the average accuracy from 45.56\% to 50.24\% in the 1-shot setting and from 62.05\% to 65.78\% in the 5-shot setting. The consistent gains indicate that incoherent regions provide more effective and challenging style variations for simulating potential target-domain shifts. By explicitly selecting and rectifying these regions, our method is able to introduce harder yet semantically guided perturbations, leading to more stable optimization and superior transferability.

\begin{figure}[!t]
\centering
\includegraphics[width=0.9\linewidth]{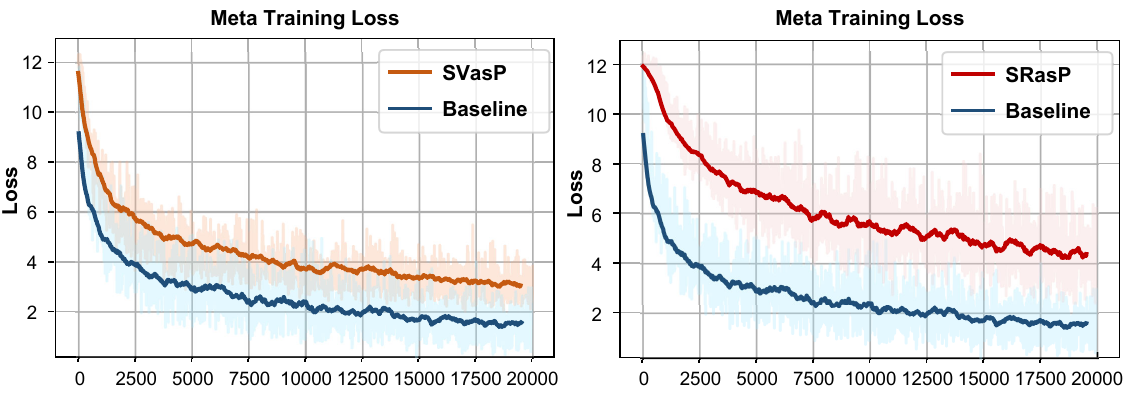}
\caption{Meta-training loss curves of the baseline and the proposed SRasP.}
\label{fig:losses_curve}
\end{figure}

\subsection{Optimization Dynamics Analysis}
As shown in the meta-training loss curves in Figure~\ref{fig:losses_curve}, SRasP consistently induces higher training losses compared to the baseline throughout the optimization process. This behavior indicates that SRasP introduces more challenging visual style perturbations during training, rather than merely acting as a regularization mechanism. Specifically, while the baseline loss quickly decreases and stabilizes, the loss under SRasP remains elevated and exhibits controlled oscillations around a higher value. Such a phenomenon suggests that SRasP actively pushes the model to confront harder and more diverse style variations, thereby increasing the difficulty of the training samples. Importantly, these oscillations remain smooth and bounded, implying that the proposed method effectively stabilizes gradient updates.

\begin{table}[t!]
  \centering
  \setlength{\tabcolsep}{1.2mm} 
  \caption{Performances on different $\xi$. The average accuracy (\%) is reported. }
    \begin{tabular}{l|l|ccc}
    \toprule
    & \textbf{$\xi$} & \textbf{BSCD-FSL} & \textbf{\textit{mini}-CUB} & \textbf{AVer.}\\
    \midrule
    \multirow{3}{*}{\textbf{1-shot}} & $\xi=0.01$ & 52.57 & 46.41 & 49.49 \\
    & \cellcolor{teal!15}{\textbf{$\xi=0.1$ (Ours)}} & \cellcolor{teal!15}{\textbf{53.26}} & \cellcolor{teal!15}{\textbf{47.22}} & \cellcolor{teal!15}{\textbf{50.24}} \\
    & $\xi=1$ & 52.68 & 46.47 & 49.58 \\
    \midrule
    \multirow{3}{*}{\textbf{5-shot}} & $\xi=0.01$ & 64.93 & 64.51 & 64.72 \\
    & \cellcolor{teal!15}{\textbf{$\xi=0.1$ (Ours)}} & \cellcolor{teal!15}{\textbf{65.92}} & \cellcolor{teal!15}{\textbf{65.65}} & \cellcolor{teal!15}{\textbf{65.78}} \\
    & $\xi=1$ & 65.11 & 64.62 & 64.86 \\
    \bottomrule
    \end{tabular}%
  \label{tab:decay}%
\end{table}%

\subsection{Parameter Sensitivity Analysis}
\subsubsection{Impact of different reorientation factors $\xi$}
Table \ref{tab:decay} reports the sensitivity of SRasP to the reorientation factor $\xi$, which controls the strength of gradient rectification for incoherent-region style perturbations. We observe that a moderate value $\xi=0.1$ consistently yields the best performance in both 1-shot and 5-shot settings across BSCD-FSL and mini-CUB. When $\xi$ is too small (\eg, $\xi=0.01$), the reorientation effect becomes insufficient, and incoherent crop style gradients are not adequately aligned with the global semantic descent direction, leading to limited performance gains. Conversely, an overly large $\xi$ (\eg, $\xi=1$) enforces excessive reorientation, which tends to over-suppress challenging style variations and slightly degrades generalization. These results indicate that an appropriate rectification strength is crucial for balancing perturbation hardness and semantic stability, and $\xi=0.1$ provides the most effective trade-off.

\begin{table}[t!]
  \centering
  \setlength{\tabcolsep}{1.8mm} 
  \caption{Performances on different $\lambda$. The average accuracy (\%) is reported. }
    \begin{tabular}{l|l|ccc}
    \toprule
    & \textbf{$\lambda$} & \textbf{BSCD-FSL} & \textbf{\textit{mini}-CUB} & \textbf{AVer.}\\
    \midrule
    \multirow{6}{*}{\textbf{1-shot}} & $\lambda=0$ & 52.77 & 46.61 & 49.69 \\
    & \cellcolor{teal!15}{\textbf{$\lambda=0.2$}} & \cellcolor{teal!15}{\textbf{53.26}} & \cellcolor{teal!15}{\textbf{47.22}} & \cellcolor{teal!15}{\textbf{50.24}} \\
    & $\lambda=0.4$ & 52.91 & 46.64 & 49.78 \\
    & $\lambda=0.6$ & 52.49 & 46.36 & 49.43 \\
    & $\lambda=0.8$ & 52.30 & 45.88 & 49.09 \\
    & $\lambda=1$ & 51.99 & 45.67 & 48.83 \\
    \midrule
    \multirow{6}{*}{\textbf{5-shot}} & $\lambda=0$ & 64.85 & 65.21 & 65.03 \\
    & \cellcolor{teal!15}{\textbf{$\lambda=0.2$}} & \cellcolor{teal!15}{\textbf{65.92}} & \cellcolor{teal!15}{\textbf{65.65}} & \cellcolor{teal!15}{\textbf{65.78}} \\
    & $\lambda=0.4$ & 65.12 & 64.77 & 64.95 \\
    & $\lambda=0.6$ & 64.96 & 64.46 & 64.71 \\
    & $\lambda=0.8$ & 64.71 & 64.27 & 64.49 \\
    & $\lambda=1$ & 64.64 & 64.37 & 64.50 \\
    \bottomrule
    \end{tabular}%
  \label{tab:lambda}%
\end{table}%

\subsubsection{Impact of different Consistency–Discrepancy Trade-off $\lambda$}
Table \ref{tab:lambda} investigates the effect of the trade-off parameter $\lambda$, which balances the consistency–discrepancy triplet objective against the standard classification loss. We observe that a moderate value of $\lambda=0.2$ achieves the best overall performance in both the 1-shot and 5-shot settings. When $\lambda=0$, the model reduces to relying solely on the classification objective, leading to inferior performance due to insufficient regulation of adversarial perturbations. As $\lambda$ increases beyond $0.2$, the performance gradually degrades. This suggests that overly emphasizing the consistency–discrepancy objective may suppress beneficial style variations and constrain feature adaptation, thereby limiting generalization. These results indicate that an appropriate balance between semantic consistency and visual discrepancy is essential, and $\lambda=0.2$ provides the most effective trade-off for robust cross-domain transfer.

\begin{table}[t!]
  \centering
  \setlength{\tabcolsep}{1.2mm} 
  \caption{Performances on whether use same $\kappa_1$, $\kappa_2$. The average accuracy (\%) is reported. }
    \begin{tabular}{l|l|ccc}
    \toprule
    & \textbf{$\kappa_1$, $\kappa_2$} & \textbf{BSCD-FSL} & \textbf{\textit{mini}-CUB} & \textbf{AVer.}\\
    \midrule
    \multirow{2}{*}{\textbf{1-shot}} & Same & 52.43 & 46.18 & 49.30 \\
    & \cellcolor{teal!15}{\textbf{Different (Ours)}} & \cellcolor{teal!15}{\textbf{53.26}} & \cellcolor{teal!15}{\textbf{47.22}} & \cellcolor{teal!15}{\textbf{50.24}} \\
    \midrule
    \multirow{2}{*}{\textbf{5-shot}} & Same & 65.01 & 64.68 & 64.85 \\
    & \cellcolor{teal!15}{\textbf{Different (Ours)}} & \cellcolor{teal!15}{\textbf{65.92}} & \cellcolor{teal!15}{\textbf{65.65}} & \cellcolor{teal!15}{\textbf{65.78}} \\
    \bottomrule
    \end{tabular}%
  \label{tab:k1_k2}%
\end{table}%

\subsubsection{Impact of different selection methods for $\kappa_1$ and $\kappa_2$}
Table \ref{tab:k1_k2} examines whether sharing the same perturbation strength for benign and adversarial branches is beneficial. Using different $\kappa_1$ and $\kappa_2$ consistently outperforms the shared setting in both the 1-shot and 5-shot scenarios. This indicates that decoupling the perturbation magnitudes provides greater flexibility to regulate benign and adversarial style variations, allowing the model to generate sufficiently hard adversarial styles while preserving stable benign representations. Consequently, the use of different $\kappa_1$ and $\kappa_2$ leads to more effective and robust cross-domain generalization.

\begin{figure}[!t]
\centering
\includegraphics[width=\linewidth]{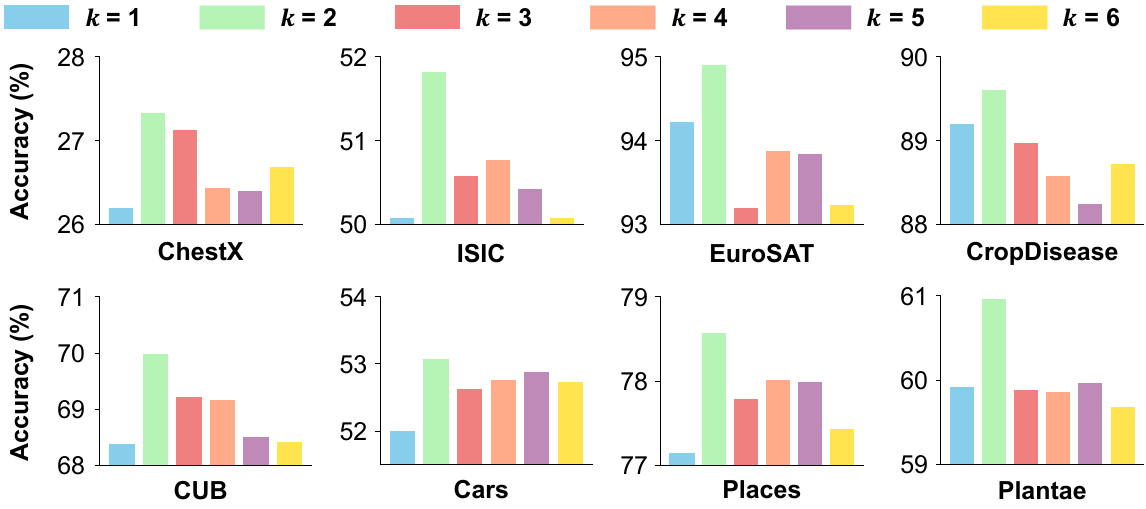}
\caption{Performances on different numbers of crops $k$.}
\label{fig:n_Crops}
\end{figure}

\subsubsection{Impact of different crop numbers $k$}
Figure \ref{fig:n_Crops} shows that incorporating multiple crops consistently improves performance over the single-crop setting across most target domains, indicating the benefit of aggregating diverse localized style information. The best results are typically achieved with a moderate number of crops, while further increasing $k$ leads to marginal gains or slight degradation due to redundant or noisy regions. These results suggest that SRasP is robust to the choice of $k$, and we therefore adopt $k=2$ as the default setting for a favorable trade-off between performance and efficiency.

\subsection{Visualization and Interpretability Analysis}
\subsubsection{Loss Landscape Analysis}
\begin{figure}[!t]
\centering
\includegraphics[width=\linewidth]{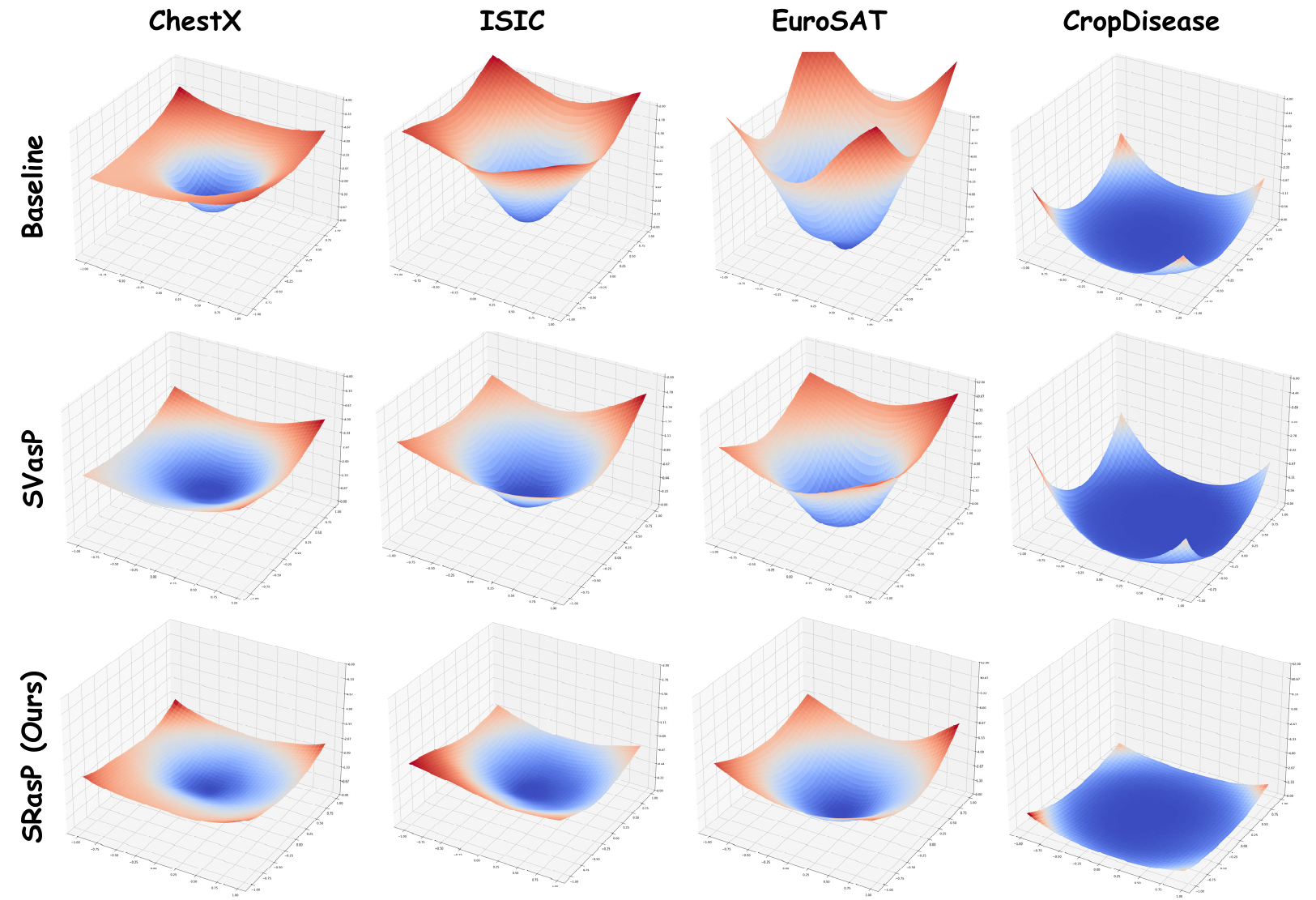}
\caption{Loss landscape visualization of different methods on the BSCD-FSL benchmark. SRasP produces markedly flatter loss landscapes across all target domains compared to the baseline and SVasP.}
\label{fig:landscape}
\end{figure}
To further investigate the optimization behavior induced by SRasP, we visualize the loss landscapes of different methods following~\cite{li2018visualizing} on the BSCD-FSL benchmark across multiple target domains, including ChestX, ISIC, EuroSAT, and CropDisease. As illustrated in Fig. \ref{fig:landscape}, the baseline model exhibits sharp and irregular loss surfaces with narrow valleys, indicating high sensitivity to parameter perturbations and a tendency to converge to unstable minima. Such sharp landscapes are particularly pronounced under severe domain shifts. The SVasP method alleviates this issue to some extent by introducing diverse adversarial style perturbations. Moreover, SRasP consistently yields flatter and smoother loss landscapes with broader basins across all evaluated domains. This indicates that the proposed SRasP method effectively suppresses incoherent-crop style gradient noise and aligns local adversarial style perturbations with the global semantic optimization direction. These observations provide strong empirical evidence that reorienting and aggregating incoherent-crop adversarial style perturbations reshapes the optimization landscape in a favorable manner.

\subsubsection{Grad-CAM Visualization Analysis}
\begin{figure}[!t]
\centering
\includegraphics[width=\linewidth]{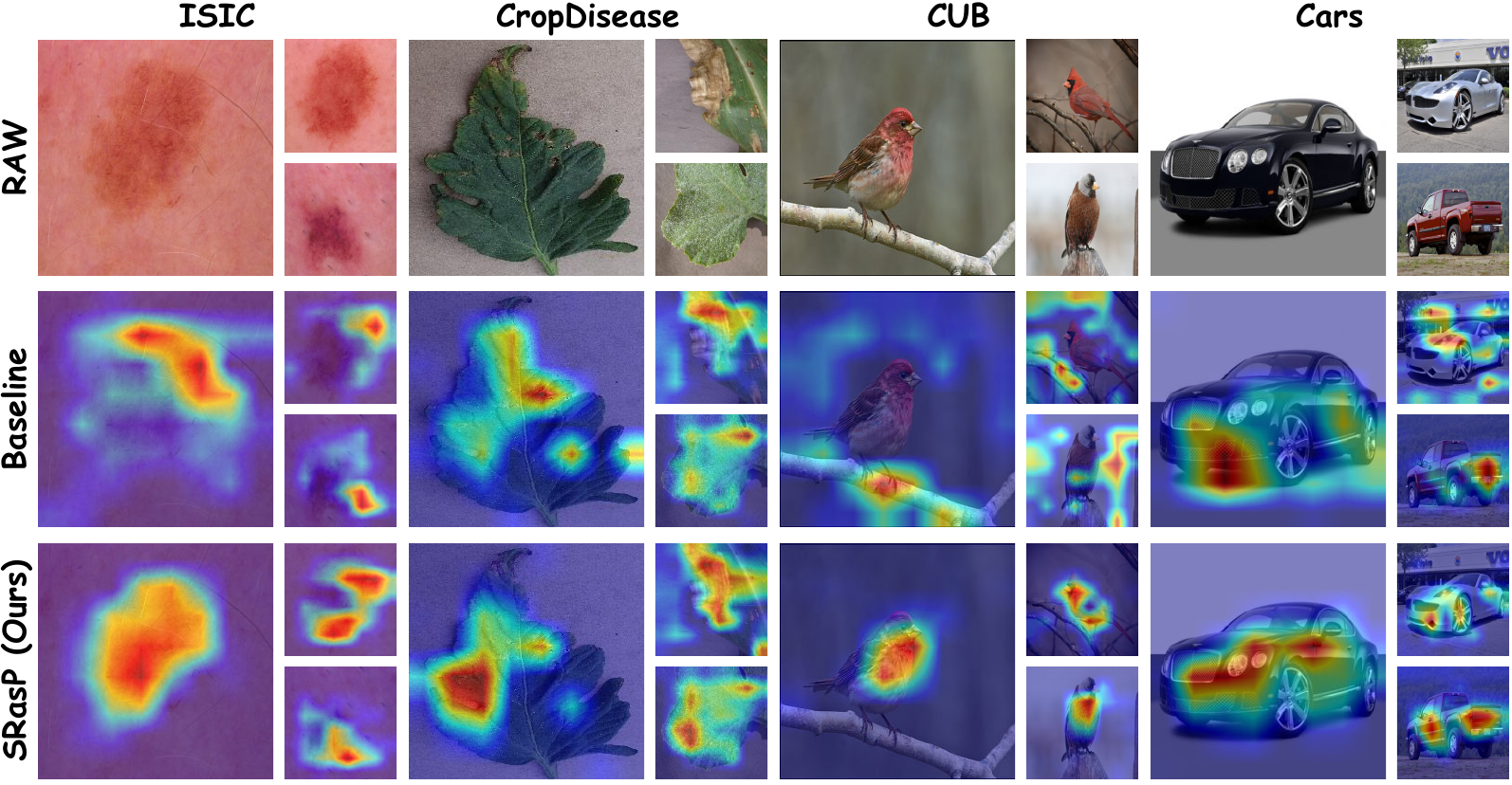}
\caption{Grad-CAM visualizations of different methods on four representative datasets. SRasP produces cleaner and more object-centric activation patterns across diverse target domains.}
\label{fig:grad_cam}
\end{figure}

To qualitatively analyze how SRasP influences the learned representations under domain shifts, we visualize Grad-CAM~\cite{selvaraju2017grad} activation maps on multiple target datasets from the BSCD-FSL benchmark. As shown in Fig. \ref{fig:grad_cam}, the baseline model frequently produces diffuse and background-biased activation patterns. In particular, the responses often extend to incoherent regions, such as homogeneous skin textures in ISIC, leaf backgrounds in CropDisease, or surrounding environments in CUB and Cars. This behavior indicates that the baseline model is prone to exploiting spurious correlations introduced by background styles, which are known to be unstable and non-transferable across domains. In contrast, SRasP consistently yields more compact, object-centric, and semantically meaningful activations. The model focuses on discriminative foreground regions, such as lesion boundaries, diseased leaf areas, bird bodies, and key vehicle parts, while substantially suppressing irrelevant background responses. By reorienting and aggregating adversarial style perturbations derived from incoherent crops, the proposed method mitigates background-induced gradient noise during training. As a result, the model learns to decouple semantic cues from background styles, leading to more reliable attention allocation at inference time. These qualitative results align well with the loss landscape analysis and quantitatively improved performance,l validating the effectiveness of SRasP in stabilizing optimization and improving generalization.

\section{Conclusion} \label{sec:conclusion}
This work focuses on style-based Cross-Domain Few-Shot Learning, addressing key challenges from the dual perspectives of regional style heterogeneity and training optimization stability. Through in-depth analysis, we reveal that adversarial style perturbations derived from incoherent crops can destabilize training when directly aggregated, leading to sharp and suboptimal minima. To address this issue, we proposed SRasP, which identifies incoherent crops under global semantic guidance and reorients their style gradients to construct stable and effective global style perturbations. By simulating hard and diverse virtual target-domain styles within each image, SRasP promotes smoother optimization trajectories and convergence to flatter, more transferable minima. Extensive experiments across multiple CD-FSL benchmarks consistently demonstrate that SRasP improves both optimization stability and cross-domain generalization. We believe that self-reoriented adversarial style perturbation offers a promising direction for robust Few-Shot Learning under severe domain shifts.

\section*{Acknowledgments}
This work was supported by the National Natural Science Foundation of China (No. 62476056 and 62306070). This work was also supported in part by the Southeast University Start-Up Grant for New Faculty under Grant 4009002309. Furthermore, the work was also supported by the Big Data Computing Center of Southeast University. This work was also supported by “the Fundamental Research Funds for the Central Universities (2242025K30024)”.

\bibliographystyle{IEEEtran}
\bibliography{reference}
\vspace{-8mm}
 \begin{IEEEbiography}[{\raisebox{0.26\height}{\includegraphics[width=0.8in,height=1in]{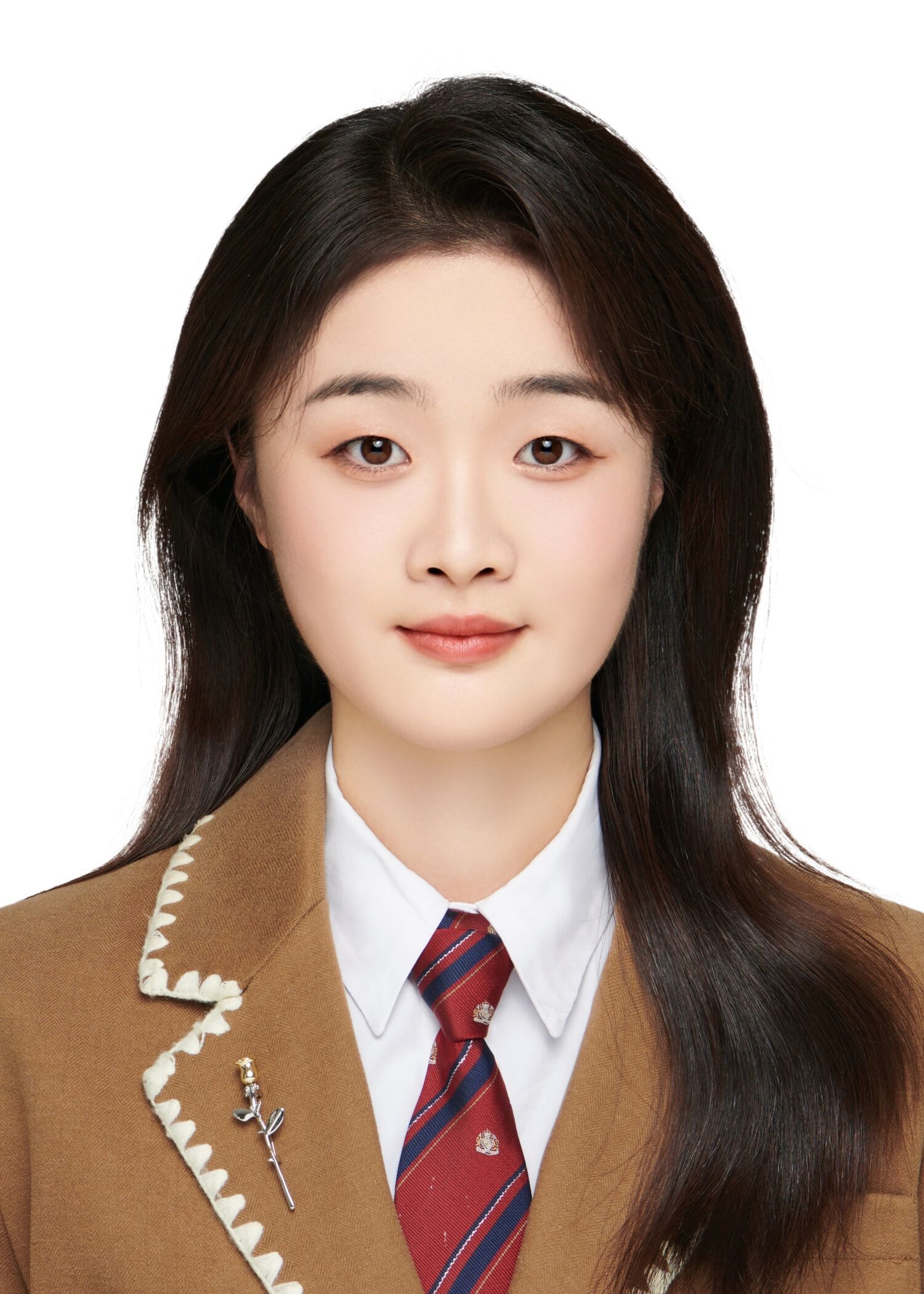}}}]{Wenqian Li}
     \scriptsize
        received the B.E. in information science and engineering technology from Southeast University in 2023. She is currently pursuing the Ph.D. degree in School of Computer Science and Engineering, Southeast University. Her research interest includes machine learning and pattern recognition.
    \end{IEEEbiography}

\vspace{-8mm}
 
\begin{IEEEbiography}[{\raisebox{0.26\height}{\includegraphics[width=0.8in,height=1in]{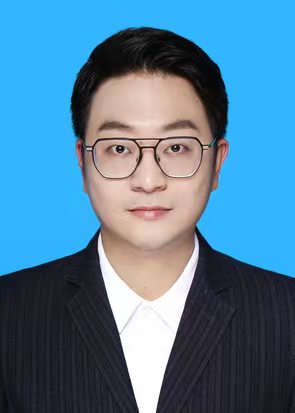}}}]{Pengfei Fang}
 \scriptsize
    is a Professor at the School of Computer Science and Engineering, Southeast University (SEU), China. Before joining SEU, he was a post-doctoral fellow at Monash University in 2022. He received the Ph.D. degree from the Australian National University and DATA61-CSIRO in 2022, and the M.E. degree from the Australian National University in 2017. His research interests include computer vision and machine learning.
\end{IEEEbiography}

\vspace{-5mm}
 
\begin{IEEEbiography}[{\raisebox{0.26\height}{\includegraphics[width=0.8in,height=1in]{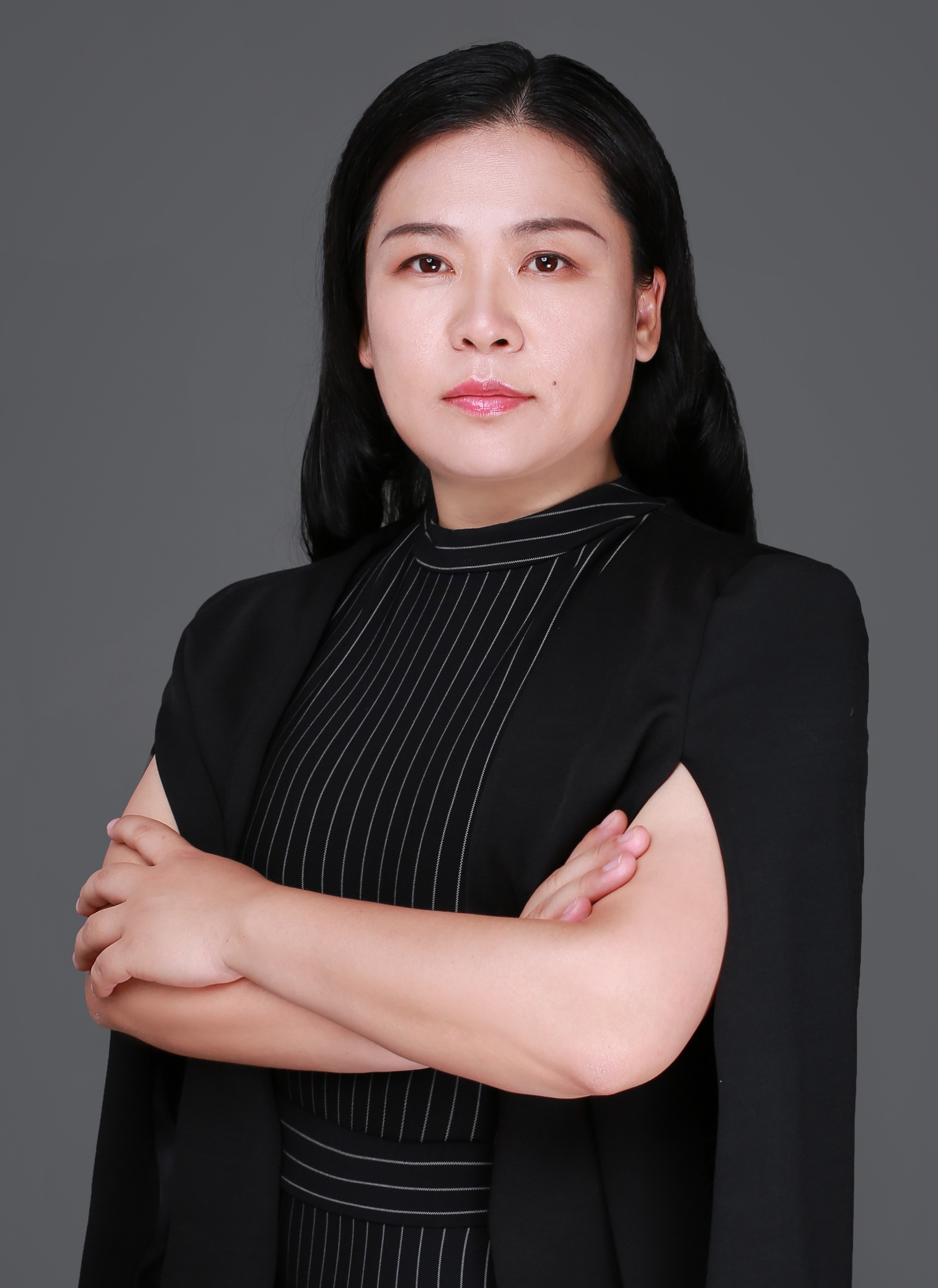}}}]{Hui Xue}(Member, IEEE)
    \scriptsize
		is currently a professor of School of Computer Science and Engineering at Southeast University, China. She received the B.Sc. degree in Mathematics from Nanjing Normal University in 2002. In 2005, she received the M.Sc. degree in Mathematics from Nanjing University of Aeronautics \& Astronautics (NUAA). And she also received the Ph.D. degree in Computer Application Technology at NUAA in 2008. Her research interests include pattern recognition and machine learning.
	\end{IEEEbiography}
    
\vfill

\end{document}